%% file: acl_latex.tex
\pdfoutput=1
\documentclass[11pt]{article}

\usepackage[final]{acl}

\usepackage{times}
\usepackage{latexsym}

\usepackage[T1]{fontenc}
\usepackage[utf8]{inputenc}

\usepackage{microtype}

\usepackage{inconsolata}

\usepackage{graphicx}
\usepackage{dsfont}

\usepackage{enumitem}
\usepackage{rotating}

\usepackage{booktabs}
\usepackage{multirow}
\usepackage{subcaption}
\usepackage{tcolorbox}
\usepackage{arydshln}
\usepackage{graphicx}
\usepackage{tabularx} 

\usepackage{amsthm}
\usepackage{amsmath}
\usepackage{amssymb}
\usepackage{xspace}
\newcommand{\narrowtextsc}[1]{\textls[-50]{\textsc{#1}}}
\newcommand{\lm}[1]{\texttt{#1}}
\newcommand{\sys}[1]{\narrowtextsc{#1}}
\newcommand{\data}[1]{\textsf{#1}}

\usepackage{xcolor,colortbl}

\definecolor{Gray}{gray}{0.94}
\definecolor{LightCyan}{rgb}{0.88,1,1}
\newcolumntype{a}{>{\columncolor{Gray}}c}
\newcolumntype{o}{>{\columncolor{white}}c}
\usepackage{soul}
\definecolor{celeste}{cmyk}{0.3922, 0.0353, 0, 0.1}
\definecolor{purple}{cmyk}{0.16, 0.28, 0, 0}
\definecolor{brilliantlavender}{cmyk}{0, 0.2235, 0, 0.1}
\definecolor{LightRed}{RGB}{232, 56, 107} 
\definecolor{LightBlue}{RGB}{116, 232, 226}
\definecolor{Tan}{rgb}{0.8203,0.7031,0.5469}
\definecolor{gblue}{RGB}{81,231,195}

\definecolor{greenblue}{RGB}{142, 207,201}

\definecolor{orange}{RGB}{255, 190, 122}

\definecolor{red}{RGB}{250, 127,111}

\definecolor{blue}{RGB}{224,230,255}

\usepackage{verbatim}

\title{\raisebox{-0.15\height}{\includegraphics[width=0.5cm]{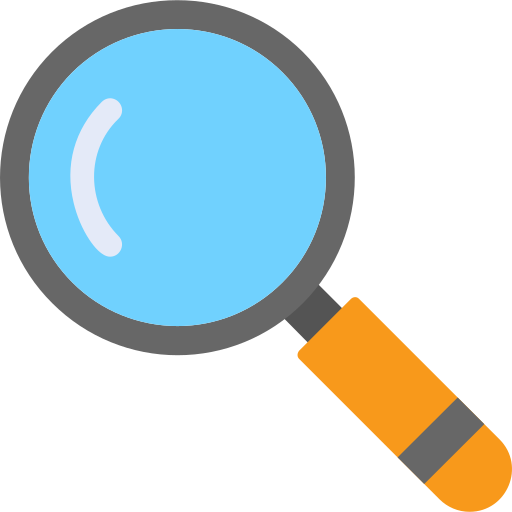}} Through a Compressed Lens: \\Investigating The Impact of Quantization on Factual Knowledge Recall}
\newcommand{\affilsup}[1]{\rlap{\textsuperscript{\normalfont#1}}}

\author{
    Qianli Wang\affilsup{1,2}
    \qquad 
    Mingyang Wang\affilsup{4,5,6}
    \qquad 
    Nils Feldhus\affilsup{1,2,8}
    \qquad
    \textbf{Simon Ostermann\affilsup{2,3,7}}
    \\
    \textbf{Yuan Cao\affilsup{9}}
    \qquad
    \textbf{Hinrich Sch\"utze\affilsup{4,6}}
    \qquad
    \textbf{Sebastian M\"oller\affilsup{1,2}}
    \qquad
    \textbf{Vera Schmitt\affilsup{1,2}}
    \\
    \small{
        $^1$Quality and Usability Lab, Technische Universit\"at Berlin
        \quad
        $^2$German Research Center for Artificial Intelligence (DFKI)
    }
    \\
     \small{
        $^3$Saarland Informatics Campus
        \quad
        $^4$LMU Munich
        \quad
        $^5$Bosch Center for Artificial Intelligence (BCAI)
    }
    \\
     \small{
        $^6$Munich Center for Machine Learning (MCML)
        \quad
        $^7$Centre for European Research in Trusted AI (CERTAIN)
    }
    \\
    \small{
        $^8$BIFOLD – Berlin Institute for the Foundations of Learning and Data
        \quad
        $^9$Technical University of Munich
    }
    \\
    \footnotesize{\textbf{Correspondence}: \texttt{\href{mailto:qianli.wang@tu-berlin.de}{qianli.wang@tu-berlin.de}}}
}

\begin{document}
\maketitle

\begin{abstract}
Quantization methods are widely used to accelerate inference and streamline the deployment of large language models (LLMs). Although quantization's effects on various LLM capabilities have been extensively studied, one critical area remains underexplored: factual knowledge recall (FKR), the process by which LLMs access stored knowledge. To this end, we conduct comprehensive experiments using three common quantization techniques at distinct bit widths, in conjunction with interpretability-driven analyses on two tasks, \textit{knowledge memorization} and \textit{latent multi-hop reasoning}.
 We show that quantization typically results in information loss within LLMs, consequently diminishing their capacity for FKR. This effect is particularly amplified in smaller models within the same architectural families. However, models quantized at reduced bit precision do not consistently exhibit inferior performance and occasionally quantization may even enhance model FKR. We find that \texttt{BitSandBytes} demonstrates highest preservation of the original full-precision model's FKR. Despite variability across models and methods, quantization causes modest performance degradation and remains an effective compression strategy. %While the effects of quantization exhibit variability across different model architectures and quantization techniques, our findings indicate no substantial performance degradation that would undermine the viability of quantization as an effective model compression strategy.

% Our code will be publicly available once the paper is published

\end{abstract}

\section{Introduction}
% \cite{gholami2022survey, treviso-etal-2023-efficient, zhu-2024-surveymodelcompressionlarge, wan2024efficient, liu-etal-2024-unlocking-data,}
% fairness \cite{ramesh-etal-2023-comparative},
% , zhu-2024-surveymodelcompressionlarge
% , mathematical reasoning \cite{liu2025quantizationhurtsreasoningempirical}
The shift towards LLMs has created strong demand for efficient inference and accessible deployment. In response, numerous quantization techniques have been developed \cite{dettmers-etal-2022-8bits, frantar-etal-2023-optq, xiao-etal-2023-smooth, lin-etal-2024-awq}. By reducing the precision of a model's parameters% and lowering the number of bits used
, quantization allows us to decrease model size while mostly preserving its performance \cite{gray-etal-1998-quantization}. Although the effects of quantization in LLMs have been evaluated across various aspects, e.g., multilinguality, bias, fairness, and trustworthiness \cite{marchisio-etal-2024-quantization, goncalves-strubell-2023-understanding, ramesh-etal-2023-comparative, hong-etal-2024-trust, liu-etal-2024-emergent, wang2026largelanguagemodelsexplain}, the \textit{degree}, \textit{variability}, and \textit{practical implications} of its impact on LLMs' ability to recall factual knowledge from pretrained memory remains underexplored and have yet to be comprehensively characterized. 

% ------------------------------------------
\begin{figure*}[t!]
\centering
\resizebox{0.95\textwidth}{!}{
\begin{minipage}{\columnwidth}
\includegraphics[width=\columnwidth]{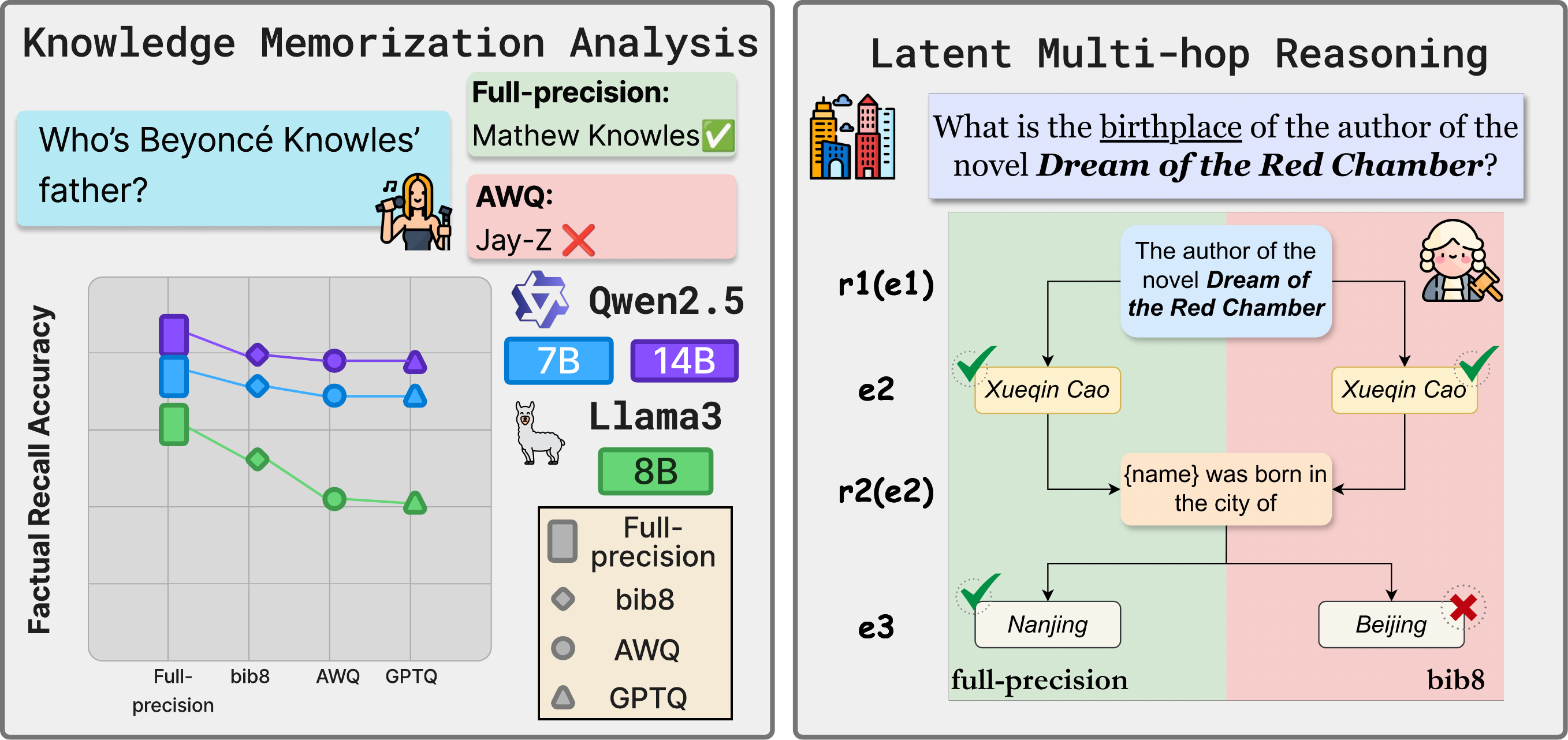}
\end{minipage}
}
\caption{The effect of quantization on factual knowledge recall through \textit{knowledge memorization analysis} and \textit{latent multi-hop reasoning analysis}.}
\label{fig:example}
\end{figure*}
% ------------------------------------------
% \cite{wang-etal-2024-unveiling}

To this end, we present a comprehensive %, interpretability-focused  
study of quantization's impact on factual knowledge recall in LLMs. Instead of solely observing the performance degradation, we perform interpretability-driven analyses on \textit{knowledge memorization} and \textit{latent multi-hop reasoning} tasks (\S\ref{subsec:methods}), to scrutinize behavior across \textbf{neuron}, \textbf{layer}, and \textbf{model} levels and evaluate three LLMs across three widely adopted model quantization techniques at different bit widths (Figure~\ref{fig:example}). We reveal that quantization typically impairs FKR, which is particularly acute for smaller models within a given model family. LLMs quantized to a lower bit precision do not consistently underperform those with higher precision. In some cases, quantization can even paradoxically enhance factual recall. Our findings indicate that \texttt{\textls[-70]BitSandBytes} is most effective at preserving the model's FKR. We observed no significant FKR degradation from quantization that would compromise model compression effectiveness, though effects vary by model and technique. %\looseness=-1

% \hl{todo} We find that quantization leads to information loss within the model, and this trend becomes more pronounced in smaller models. Moreover, larger quantized models do not consistently outperform smaller full-precision models and quantization-induced degradation on explanation quality depends on the specific quantization techniques, and models deployed. Generally, integer quantization \texttt{bib8} yields superior explanation quality compared to other quantization techniques across the evaluated methods. Surprisingly, lower-precision LLMs can sometimes surpass their higher-precision counterparts and a quantized model may even generate better explanations than the same model in full precision. 

\section{Background and Related Work}
\label{sec:background}
\paragraph{Quantization.} 
% Quantization has been demonstrated to offer a superior balance between compression and accuracy \cite{li2024evaluating}. 
%The decoding stage during LLM inference is typically memory-bound, where the key-value cache (KV cache) overhead often exceeds the size of the model weights \cite{li2024evaluating}. 
Quantization techniques compress LLMs by converting model weights, activations, or the KV cache into lower-precision data types \cite{zhu-2024-surveymodelcompressionlarge}. These techniques can be broadly categorized into two types: quantization-aware training (QAT) and post-training quantization (PTQ). QAT requires retraining to mitigate errors introduced by quantization, whereas PTQ facilitates the direct application of a quantized model during inference. In this paper, we primarily evaluate the impact of weight-only quantization (\S\ref{subsec:quantization}) on FKR (\S\ref{sec:evaluation}), which eliminates the need for retraining to address errors resulting from quantization.
 % PTQ addresses this by reducing the memory usage of weights, activations, and KV cache using low-precision values \cite{zhu-2024-surveymodelcompressionlarge}.

\paragraph{Impact of Quantization.}
Recent work has extensively examined the impact of quantization on various capabilities of LLMs. \citet{marchisio-etal-2024-quantization} conduct a thorough analysis of quantized multilingual LLMs, focusing on performance degradation across languages. \citet{goncalves-strubell-2023-understanding, kirsten-etal-2024-bias} explore the emergence of bias in the outputs generated by quantized models. \citet{liu-etal-2024-emergent} find that in-context learning ability gradually declines in heavily quantized LLMs. \citet{jin-etal-2024-comprehensive} observe that models with 4-bit quantization can still retain the alignment ability. \citet{singh-sajjad-2025-interpreting} investigate the impact of quantization on model calibration. \citet{wang2026largelanguagemodelsexplain} explore the impact of quantization on self-explanation quality and faithfulness. In our work, we explicitly explore how quantization affects FKR.\looseness=-1

\section{Experimental Setup}
We examine two representative methods (\S\ref{subsec:methods}) to evaluate the impact of quantization on FKR. Specifically, we compare full-precision LLMs (\S\ref{subsec:models}) with LLMs quantized using different techniques and bit configurations (\S\ref{subsec:quantization}) across two datasets (\S\ref{subsec:datasets}).

\subsection{Methods}
\label{subsec:methods}

\paragraph{Knowledge Memorization Analysis.}
%This investigation focuses on understanding the effects of quantization on the 
% As a first interpretability technique, 
We investigate the memorization of a model's factual knowledge, identifying the reasons behind potential factual forgetting \cite{namburi-etal-2023-cost}. By leveraging the theory of knowledge neurons, which suggests that specific neurons in LLMs are responsible for storing specific pieces of knowledge, we explore how quantization alters the storage and retrieval processes within these neurons. %Our findings aim to shed light on the trade-offs between model efficiency and fidelity to factual information, providing deeper insights into the impact of quantization on the internal mechanisms of LLMs.

\paragraph{Latent Multi-hop Reasoning Analysis.}

We adapt the methodology from \citet{yang-2024-latent-multi-hop-reasoning} for inspecting latent multi-hop reasoning errors. Specifically, the analysis tests whether full-precision and quantized LLMs employ similar latent reasoning pathways and internal knowledge recall mechanisms to answer complex factual queries.
%Since the authors reported a clear scaling trend for the first hop of reasoning but not for the second hop, we aim to investigate if this pattern can also be found for full-precision and quantized LLMs.

%and \texttt{LLM.int8()} implemented by \sys{BitsAndBytes}\footnote{Although \texttt{LLM.int8()} belongs to weight-activation quantization, the implementation provided in \sys{BitsAndBytes} is limited to weight-only quantization.} \cite{dettmers-etal-2022-8bits, dettmers-etal-2023-qlora}.

\subsection{Datasets}
\label{subsec:datasets}
Our study employs two widely recognized datasets\footnote{Dataset examples are detailed in Appendix~\ref{app:dataset}.} for evaluating FKR with selected interpretability methods (\S\ref{subsec:methods}).

\paragraph{LRE} 
\cite{hernandez-2024-linearity} is a knowledge probing dataset consisting of knowledge triplets $(s, r, o)$, structured in a one-hop setting, where $s$, $r$, and $o$ represents the subject, relation, and object, respectively.

\paragraph{TwoHop-Fact} \cite{yang-2024-latent-multi-hop-reasoning} is a dataset consisting of pairs of prompts: two-hop prompts ($\tau_{2H}$) for compositional queries, representing fact composition queries in the form $((e_1,r_1,e_2),(e_2,r_2,e_3))$, where $r_1$ and $r_2$ are relations and $e_i$ denotes an entity; and one-hop prompts ($\tau_{1H}$) for subqueries (Figure~\ref{fig:TwoHopFact_example}). $e_1$ and $e_2$ (in the second triplet) are subjects, while $e_2$ (in the first triplet) and $e_3$ objects. $e_2$ serves as a \textbf{bridge entity}, linking the two triplet to form a coherent two-hop reasoning chain. 

% An exemplary query consists of $s \leftarrow \text{"\textit{the company that created Visual Basic}"}$ and $o \leftarrow \text{"\textit{The current CEO of}"}$. A correct answer (in this example: \textit{Satya Nadella}) by the explained model is the criterion by which the LRE data is filtered.

\subsection{Quantization Techniques}
\label{subsec:quantization} Building on the prior discussion (\S\ref{sec:background}), we identify three commonly used PTQ techniques applied to the selected LLMs (\S\ref{subsec:models}) in our experiments: \looseness=-1
% [noitemsep,topsep=0pt,leftmargin=*]
\begin{itemize}[leftmargin=*]
    \item \texttt{GPTQ} \cite{frantar-etal-2023-optq} uses a second-order, Hessian-based optimization to quantize weights post-training with minimal accuracy loss;
    \item \texttt{AWQ} \cite{lin-etal-2024-awq} enhances weight quantization by handling activation outliers to preserve model accuracy at low bit-widths; 
    \item Integer quantization \cite{dettmers-etal-2022-8bits} implemented by \sys{BitsAndBytes} (\texttt{bib4} and \texttt{bib8}) enables fast and memory-efficient inference by using optimized low-bit kernels. \looseness=-1
\end{itemize}
% \footnote{The implementation of integer quantization provided in \sys{BitsAndBytes} is limited to weight-only quantization: \url{https://github.com/bitsandbytes-foundation/bitsandbytes}.}

\subsection{Models}
\label{subsec:models}
We evaluate \lm{Llama3-8B} \cite{llama3modelcard}, \lm{Qwen2.5-7B}, and \lm{Qwen2.5-14B} \cite{qwen2024qwen25technicalreport} using two interpretability methods (\S\ref{subsec:methods}). These models are selected due to the availability of quantized versions for each\footnote{Detailed information for each model, including links to their respective quantized versions, is provided in Table~\ref{tab:used_model}.}, ensuring the reproducibility of our results.

\begin{figure*}[t!]
    \centering
    \begin{minipage}{\textwidth}
        % \centerline{\textbf{\lm{Qwen2.5-7B}}}
        \vspace{0.1cm}
                \begin{minipage}{0.48\textwidth}
            \centering
            \includegraphics[width=0.85\textwidth]{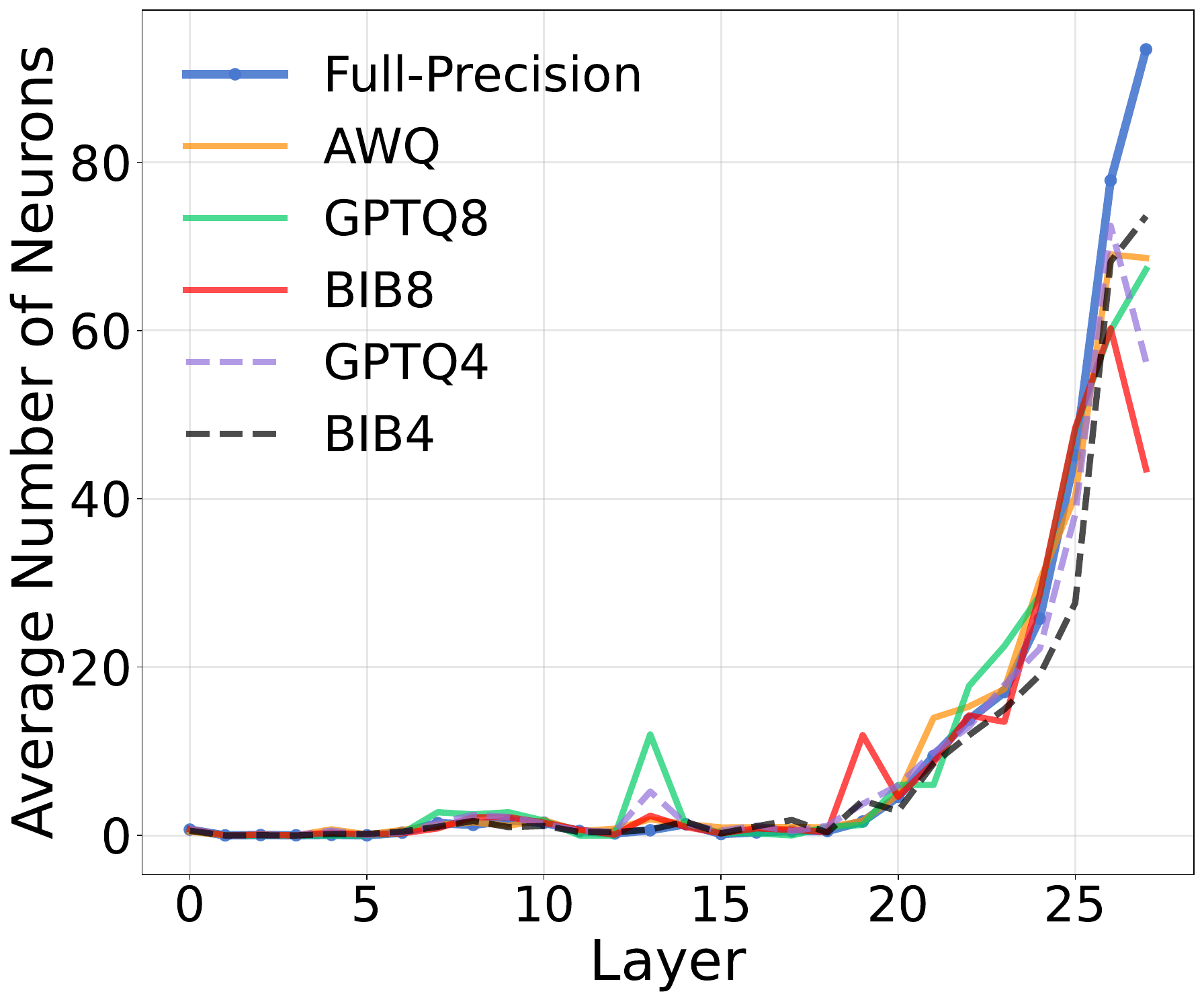}
            \subcaption{Top neuron distribution (\lm{Qwen2.5-7B})}
            \label{fig:neuron-dist-qwen}
        \end{minipage}%
        \hfill%
        \begin{minipage}{0.48\textwidth}
            \centering
            \includegraphics[width=0.85\textwidth]{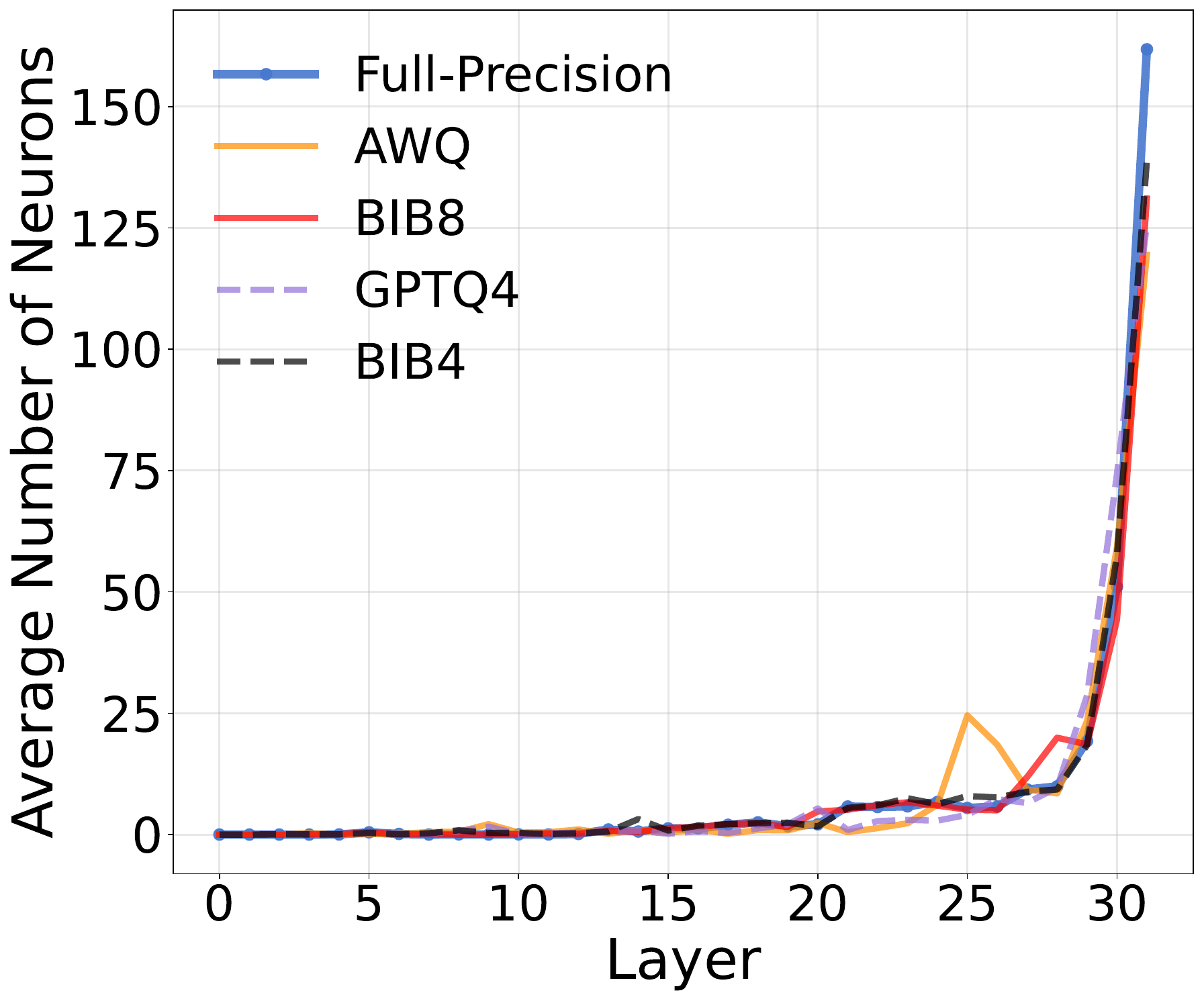}
            \subcaption{Top neuron distribution (\lm{Llama3-8B})
            }
            \label{fig:neuron-dist-llama}
        \end{minipage}
   
        \vspace{0.2cm}
        % \centerline{\textbf{\lm{Llama3-8B}}}
        \vspace{0.4cm}
     \begin{minipage}{0.49\textwidth}
            \centering
            \includegraphics[width=0.85\textwidth]{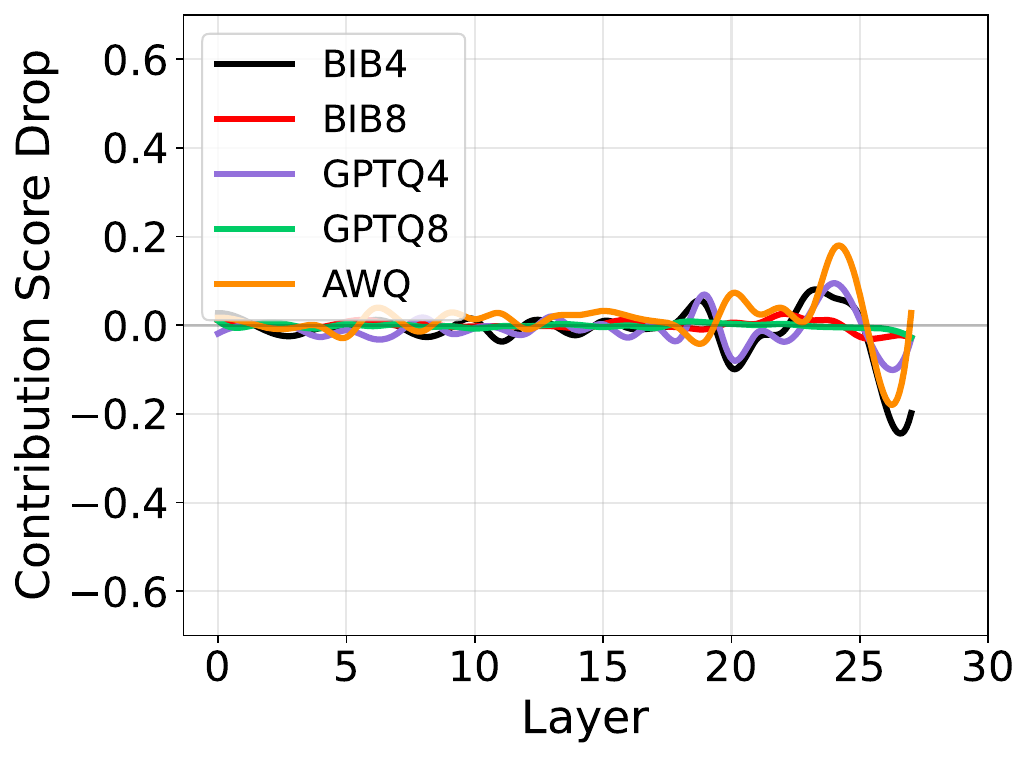}
            \label{fig:layer-attn}
            \subcaption{Attention (\lm{Qwen2.5-7B}): Landmark on continent}
        \end{minipage}%
        \hfill%
        \begin{minipage}{0.49\textwidth}
            \centering
            \includegraphics[width=0.85\textwidth]{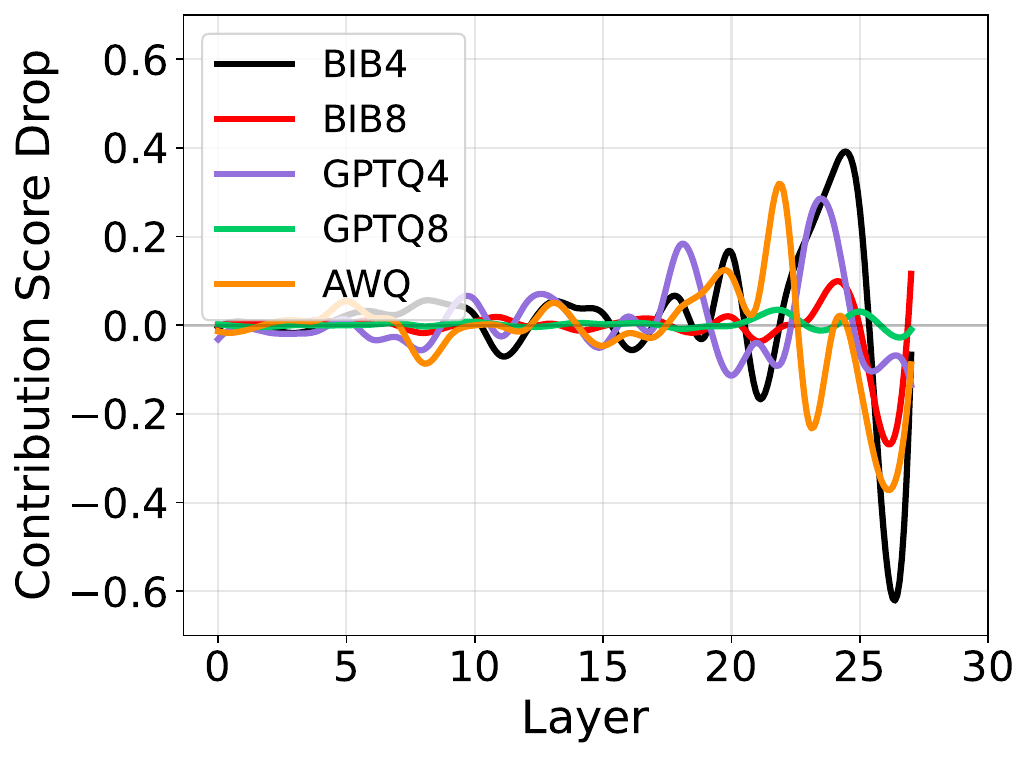}
            \label{fig:layer-ffn}
            \subcaption{FFN (\lm{Qwen2.5-7B}): Landmark on continent}
        \end{minipage}
    \end{minipage}
    
    \caption{\textbf{Top:} Distribution of high-contributing neurons across layers, showing the average number of top-300 neurons per layer for \lm{Qwen2.5-7B} (\textit{left}) and \lm{Llama3-8B} (\textit{right}). \textbf{Bottom:} Layer-wise drop in neuron contribution scores across quantization methods for the \textit{landmark on continent} relation in \lm{Qwen2.5-7B}, comparing attention sublayers (\textit{left}) and feed-forward sublayers (\textit{right}).}
    \label{fig:layer_contribution_comparison}
\end{figure*}

\section{Evaluation Setup}
\label{sec:evaluation}

\paragraph{Knowledge Memorization Analysis.} To assess the effects of quantization on FKR, we first evaluate factual recall accuracy using \data{LRE}. A comparison of results before and after quantization reveals the degree of knowledge forgetting.
We then employ a knowledge attribution method \citep{yu-ananiadou-2024-neuron} to trace the information loss back to specific layers and neurons, following the knowledge neuron theory. This combined analysis uncovers how quantization impacts the internal mechanisms responsible for storing and retrieving information.
% retrieving factual information

\paragraph{Latent Multi-hop Reasoning Analysis (LMHR).} Following \citet{yang-2024-latent-multi-hop-reasoning}, we employ three metrics to evaluate the impact of quantization on FKR, i.e., LMHR. The Entity Recall Score (\sys{EntRec}) measures the LLM's ability to recall the bridge entity $e_2$ within a two-hop prompt $\tau_{2H}$ (Figure~\ref{fig:TwoHopFact_example}). \sys{EntRec} is defined with respect to the hidden representation in a specific layer $\ell$, at the final position of the bridge entity’s descriptive mention in the two-hop prompt. A higher \sys{EntRec}$_{\ell}(e_2,\tau_{2H})$ indicates stronger internal recall of the bridge entity $e_2$ at the $\ell$-th layer.
The Consistency Score (\sys{CnstScore}) assesses how consistently an LLM responds to both the two-hop and one-hop prompts. 
\sys{CnstScore} calculates the similarity between the output probability distributions in response to the $\tau_{2H}$ and $\tau_{1H}$ prompts to measure the consistency between the two outputs.
Additionally, we evaluate FKR accuracy in predicting the target object $e_3$. \looseness=-1

\section{Results}

\subsection{Knowledge Memorization Analysis}
\label{subsec:knowlegde_results}
% \textbf{1. Knowledge Recall Performance Drop.}
Table~\ref{tab:lre_acc} shows that quantization introduces varying degrees of accuracy drop depending on both the model size and quantization method. Notably, the accuracy drop is more obvious in smaller models within the same model family. Additionally, the accuracy of \texttt{AWQ}, \texttt{GPTQ4}, and \texttt{bib4} quantized models consistently drop across all model sizes, whereas \texttt{GPTQ8} and \texttt{bib8} models retain performance comparable to the full-precision models. %This effect is especially pronounced in \lm{Qwen2.5-\{7,14\}B}. 
% Furthermore, the factual recall accuracy degradation is similar across different relations irrespective of whether the full-precision performance has saturated or not (Table~\ref{tab:per-relation-factual-recall-acc-7B}, Table~\ref{tab:per-relation-factual-recall-acc-14B}). This indicates that the factual knowledge exhibits similar fragility under quantization. 
Furthermore, relations where full-precision performance has not saturated tend to exhibit more severe factual knowledge recall degradation (Table~\ref{tab:per-relation-factual-recall-acc-7B}), indicating that such knowledge is more fragile under quantization.
% Furthermore, the factual recall accuracy degradation is more severe in relations where the full-precision performance has not yet saturated (Table~\ref{tab:per-relation-factual-recall-acc-7B}), indicating that such knowledge is more fragile under quantization. %We provide detailed per-relation accuracy breakdowns in the appendix to further illustrate this trend.

\paragraph{Neuron-level Trends.} Following the neuron-level method of \citet{yu-ananiadou-2024-neuron}, we assign each neuron in the model a contribution score equal to the increase in log-probability of the correct answer token that the neuron induces. Comparing these scores before and after quantization allows us to quantify how much individual neurons' influence changes. We analyze four relations that suffer a notable accuracy drop: \textit{landmark on continent}, \textit{person father}, \textit{person mother}, and \textit{person sport position}.

We track the top-300 feed-forward neurons in full-precision model, set $\tau$ as their minimum contribution score, and count how many neurons in quantized models exceed $\tau$. As shown in Figure \ref{fig:neuron-dist-qwen} and \ref{fig:neuron-dist-llama}, the per-layer counts decrease after quantization for both models, with the most obvious reductions in the last layers. Complete results across relations are provided in Figures~\ref{fig:landmark_continent_analysis} and \ref{fig:landmark_continent_analysis_llama}
% ; additional relations are reported 
in App.~\ref{app:memorization}. 

\paragraph{Layer-wise Trends.} Figure \ref{fig:layer_contribution_comparison} depicts the drop in aggregate contribution scores for attention (\textit{left}) and feed-forward (\textit{right}) sub-layers. All quantization methods exhibit a pronounced decline in the final two layers on \lm{Qwen2.5-7B}, this trend is consistent on all relations we investigate, as shown in Figure~\ref{fig:layer_contribution_comparison_all} in App~\ref{app:memorization}. While on \lm{Llama3-8B} (see Figure~\ref{fig:layer_contribution_comparison_all_llama}), the decline occurs in middle-to-late layers, with an increase in the final layers, indicating that information loss patterns vary across model architectures. This divergence may be ascribable to the different ways in which factual knowledge is stored across model families \cite{choe-etal-2025-autoregressive}.
 % The layer-level findings align with the neuron-level observations that most information loss occurs during the network’s decision stages.

% \input{table/LMHR}

% \textbf{Neuron-wise Trends.} We further track the 300 feed-forward neurons with the highest contribution scores in full-precision model. 
% As shown in Figure \ref{fig:neuron-dist-qwen} and \ref{fig:neuron-dist-llama}, the per-layer counts decrease after quantization for both models, with the most obvious reductions in the last layers. These neuron-level findings algin with the layer-level observation described above.

% Collectively, these findings confirm that quantization diminishes decisive information stored in the network’s late layers, leading to the factual knowledge recall degradation reported in Table \ref{tab:lre_acc}.

\paragraph{Summary.} Collectively, both layer-level and neuron-level analyses reveal that quantization primarily affects the network's last layers. These findings confirm that quantization degrades the decisive information stored in these late layers, accounting for the factual knowledge recall degradation reported in Table \ref{tab:lre_acc}.\footnote{The complete analysis of both models across different relations is provided in Appendix~\ref{app:memorization}.}

\begin{figure*}[t]
    \centering
    \begin{subfigure}[b]{0.8\textwidth}
        \centering
        \includegraphics[width=\textwidth]{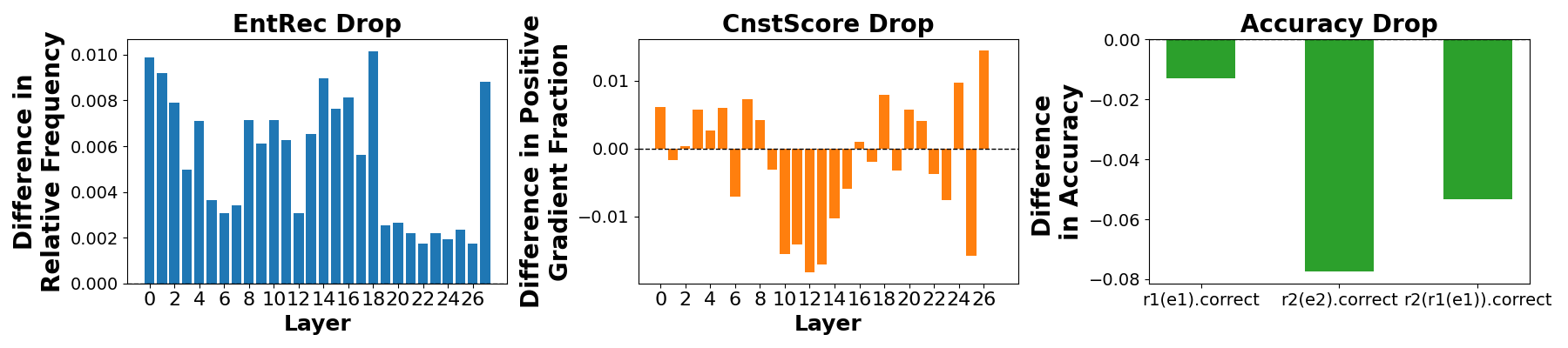}
        \caption{\lm{Qwen2.5-7B}}
        % \label{fig:subfig1}
    \end{subfigure}

    \begin{subfigure}[b]{0.8\textwidth}
        \centering
        \includegraphics[width=\textwidth]{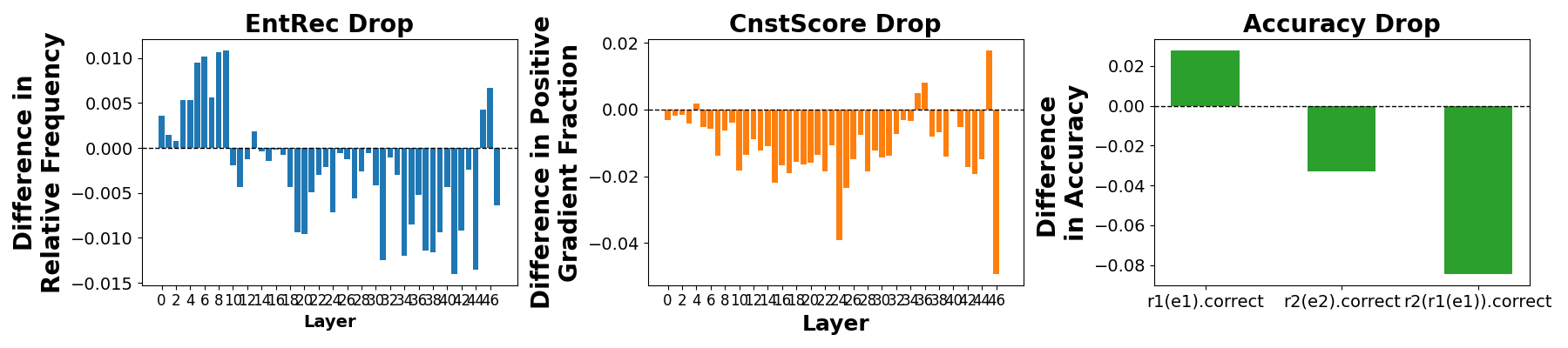}
        \caption{\lm{Qwen2.5-14B}}
        % \label{fig:subfig1}
    \end{subfigure}

    \begin{subfigure}[b]{0.8\textwidth}
        \centering
        \includegraphics[width=\textwidth]{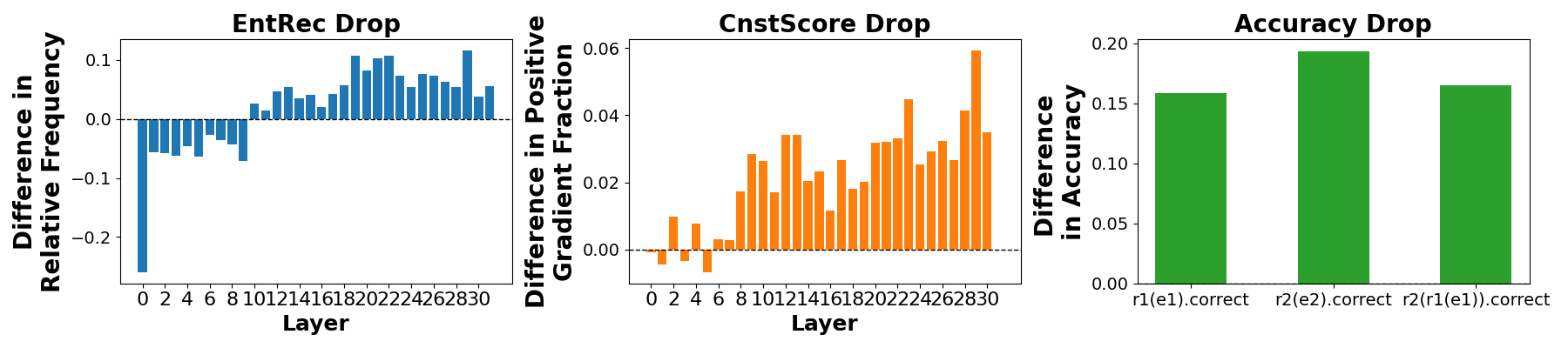}
        \caption{\lm{Llama3-8B}}
        \label{subfig:llama3}
    \end{subfigure}
    
    \caption{Difference in the \textit{entity recall score} (\sys{EntRec}), \textit{consistency score} (\sys{CnstScore}), and \textit{accuracy} between the \texttt{GPTQ8} quantized and full-precision models, evaluated across all layers.}
    \label{fig:lmhr_diff}
\end{figure*}

\subsection{Latent Multi-hop Reasoning Analysis}
\label{subsubsec:lmhr_results}

%Table~\ref{tab:lmhr} reveals that LLMs generally perform better on subqueries, i.e., $r_1(e_1)$ and $r_2(e_2)$, compared to complex queries that require a compositional reasoning chain $r_2(r_1(e_1))$, suggesting challenges in composing multiple reasoning steps. 

% Figure~\ref{fig:lmhr_diff} outlines that, for \lm{Qwen2.5-\{7B,14B\}}, quantization generally leads to weaker internal recall of the bridge entity $e_2$ across most layers, while the degradation of output $e_3$ consistency across layers remains mixed. 

% Table~\ref{tab:lmhr_llama} reveals that \texttt{bib} is the most effective quantization method at preserving FKR. Meanwhile,

\paragraph{Quantization affects the first-hop Reasoning the most.} Table~\ref{tab:lmhr_llama} reveals that quantization substantially affects the first hop $r_1(e_1)$, by as much as $30.08\%$, while its impact on the second hop is minimal, with an average of degradation of only $4.25\%$. The $r_2(r_1(e_1))$ deterioration due to quantization is strongly correlated with the ability to correctly predict the bridge entities $r_1(e_1)$, as indicated by a Spearman's correlation of $0.93$. Nevertheless, the FKR deterioration is not dramatic and considerably acceptable (e.g., \lm{Qwen2.5-7B} shows a minor average deterioration of 0.77\%). \looseness=-1

\paragraph{Quantization effects are not consistent.} Figure~\ref{fig:lmhr_diff} illustrates that quantization effects on FKR are largely unpredictable. The quantization effect is heterogeneous across layers and variable across models, particularly across architectures, given that facts are stored in various ways across \lm{Llama3} and \lm{Qwen2.5} models \cite{choe-etal-2025-autoregressive}. Besides, different quantization approaches often affect the model FKR in different manners (Figure~\ref{fig:diff_7b}). Nevertheless, for a given quantization method with different bit width, the layer-wise impact remains broadly similar. Surprisingly, quantization can occasionally even improve FKR (Figure~\ref{fig:lmhr_diff}). This effect may be attributed to the regularization effect \cite{park-etal-2022-robust} or quantization-induced noise \cite{li2024investigating}, which can inadvertently enhance the model's capability to recall factual knowledge. 

\paragraph{\texttt{bib} largely preserve the factual knowledge recall capability and cross-model comparison.} For \lm{Qwen2.5-7B}, quantization consistently reduce knowledge recall, with the largest degradations under \texttt{GPTQ} and \texttt{AWQ} (Figure~\ref{fig:diff_7b}). By contrast, \texttt{bib4/8} degrades one-hop reasoning $r_1(e_1)$ and $r_2(e_2)$, as reflected by modest shifts in \sys{EntRec} and \sys{CnstScore} relative to the full-precision model, but leaves two-hop reasoning largely unaffected. For \lm{Qwen2.5-14B}, \texttt{bib4/8} likewise outperforms other quantization methods, best preserving and occasionally improving FKR (Figure~\ref{fig:diff_14b}). Moreover, despite its greater layer depth than \lm{Qwen2.5-7B}, \lm{Qwen2.5-14B} exhibits a similar layer-wise pattern of \texttt{bib4/8} quantization effects (Figure~\ref{fig:7b_bib4}, \ref{fig:14b_bib4}), whereas other quantization methods do not.  For \lm{Llama3-8B}, the effect of different quantization methods are alike, i.e., \sys{EntRec} and \sys{CnstScore} of quantized model are lower than full-precision model in shallow layers, but becomes much higher in deeper layers (Figure~\ref{fig:diff_llama}).  Among the three LLMs evaluated, \lm{Llama3-8B} is least affected by quantization in terms of FKR.

% \subsection{Discussion} 
% In general, quantization leads to information loss within the model, which in turn degrades knowledge recall (Table~\ref{tab:lre_acc}, Table~\ref{tab:lmhr_llama}); this trend becomes more pronounced for smaller models within the same family (Figure~\ref{fig:diff_7b}, Figure~\ref{fig:diff_14b}). Moreover, LLMs quantized with lower bit precision do \textit{not} invariably perform worse than those with higher bit precision. Quantization can occasionally even improve FKR (Figure~\ref{fig:lmhr_diff}). This effect may be attributed to the regularization effect \cite{park-etal-2022-robust} or quantization-induced noise \cite{li2024investigating}, which can inadvertently enhance the model's capability to recall factual knowledge. Overall, across the selected use cases, \texttt{bib} most closely preserves the full-precision model’s FKR. Additionally, quantization's impact differs across models and methods, but we found no significant degradation that would compromise its effectiveness for model compression.

\section{Conclusion}
In this work, we examined the impact of quantization on factual knowledge recall. %, evaluated on two datasets using three quantization techniques across three LLMs. 
In general, quantization leads to information loss within the model, which in turn degrades factual knowledge recall; this trend becomes more pronounced for smaller models within each model family. Moreover, LLMs quantized with lower bit precision do \textit{not} invariably perform worse than those with higher bit precision. Quantization can occasionally even improve factual knowledge recall. %We find that \texttt{bib4/8} most closely preserves the model’s FKR. 
While quantization effects vary by model and technique, we observed no performance degradation severe enough to compromise its viability as a compression strategy. \looseness=-1
% While the effects of quantization vary depending on the specific model and technique employed, we observed no performance degradation substantial as to compromise its effectiveness as a viable strategy for model compression.

% In general, quantization leads to information loss within the model, and this effect becomes more pronounced in smaller models. The extent of explanation quality degradation is dependent on the specific quantization methods and the models employed. Lower precision LLMs can sometimes outperform their higher-precision counterparts, and quantized models may even surpass full-precision models, particularly in the case of larger LLMs. Overall, integer quantization (\texttt{bib8}) yields higher-quality explanations compared to other quantization techniques across the selected interpretability methods. 

% \clearpage

\section*{Limitations}
Our experimental work is confined to English-language datasets. Consequently, the effectiveness of our experiments in other languages may not be comparable and multilingual factual knowledge recall may be simultaneously affected by the degradation of multilingual capabilities due to quantization \cite{marchisio-etal-2024-quantization}. Extending experiments to the multilingual settings is considered for future work.

We restrict our experiments to \lm{Llama3-8B}, \lm{Qwen2.5-7B} and \lm{Qwen2.5-14B} for computational feasibility: our gradient-based analyses exceed the available GPU memory for larger models. Between the model families (Qwen, Llama, Mistral, DeepSeek), there are lots of architectural equivalences and similarities, e.g., the same attention (grouped-query attention), position embeddings (RoPE), normalization (RMSNorm) or FFN activation (SwiGLU). We argue that, based on our comprehensive experiments, our results are considerably robust and generalizable given the similar architectures compared to other model families. 

In our experiments, we extensively compare full-precision models with different quantized versions in 4-bit and 8-bit formats. Lower-bit quantization, such as 1-bit or 2-bit, is not included in our study. 

Moreover, the scope of our experiments is limited to post-training quantization techniques. Investigating the impact of weight-activation quantization, KV cache compression, or quantization-aware training techniques on factual knowledge recall is counted as future work.

\section*{Author Contributions}
Author contributions are listed according to the CRediT taxonomy as follows:
\begin{itemize}[noitemsep,topsep=0pt,leftmargin=*]
    \item QW: Writing, idea conceptualization, experiments and evaluations, analysis, visualization.
    \item MW: Writing, experiments and evaluations, and analysis.
    \item NF: Writing – review \& editing, experiments and evaluations, and supervision.
    \item SO: Writing – review \& editing.
    \item YC: Experiments and evaluations.
    \item HS: Supervision.
    \item SM: Supervision and funding acquisition.
    \item VS: Funding acquisition and proof reading.
\end{itemize}

\section*{Acknowledgment} 
We are indebted to the anonymous reviewers of TrustNLP workshop co-located at ACL 2026 for their helpful and rigorous feedback.
This work has been supported by the Federal Ministry of Research, Technology and Space (BMFTR) as part of the project VERANDA (16KIS2047).

\bibliography{custom}
% \bibliography{anthology,custom}

% \clearpage

\appendix
\section{Dataset Information}
\label{app:dataset}

\paragraph{LRE} An exemplary query consists of $s \leftarrow \text{"\textit{the company that created Visual Basic}"}$ and $o \leftarrow \text{"\textit{The current CEO of}"}$. A correct answer (in this example: \textit{Satya Nadella}) by the explained model is the criterion by which the \data{LRE} data is filtered.

\paragraph{TwoHop-Fact} Figure~\ref{fig:TwoHopFact_example} illustrates an example from the \data{TwoHop-Fact} dataset. Based on the input text, $(e_1, r_1, e_2)$ corresponds (Superstition, singer, Stevie Wonder), while $(e_2, r_2, e_3)$ represents (Stevie Wonder, mother, Lula).

\begin{figure}[t!]
    
    \centering

    \begin{tcolorbox}[colback=pink!20!white, colframe=pink!80!black, title=TwoHop-Fact (Multi-hop Reasoning)]
    \textbf{One-hop prompts}: 
    ($p_1$): The mother of Stevie Wonder is Lula. 
    ($p_2$): The singer of `Superstition' is Stevie
    Wonder.
    \bigskip
    
    \textbf{Two-hop prompt}: 
    ($p$): The mother of the singer of `Superstition'
    is
    \bigskip
    
    \textbf{Answer}: Lura
    \end{tcolorbox}
    
    \caption{An example from the \data{TwoHop-Fact} dataset.
    }
    \label{fig:TwoHopFact_example}
\end{figure}

\section{Models \& Inference Time}
\input{table/model}
Table~\ref{tab:used_model} presents details of the all LLMs used in our experiments (\S\ref{subsec:models}), including model sizes, quantization approaches and corresponding URLs from the Hugging Face Hub. All models were directly obtained from the Hugging Face repository. All experiments were conducted using A100 or H100 GPUs. Neuron-level and layer-level attribution can be completed within 10 hours, while LMHR experiments take 30 hours averagely.

% \section{Prompt Instructions}

% \section{LLM-as-a-Judge}
% \label{app:llm_as_a_judge}
% Following \citet{gabryszak-etal-2024-enhancing-editorial}, we deploy \lm{GPT-4o} as the judge and use the prompts shown in Figure~\ref{fig:llm_as_a_judge_prompt} to evaluate the generated counterfactual examples and natural language explanations based on trustworthiness, coherence, and insightfulness.

% % ------------------------------------------
% \begin{figure*}[t!]
% \centering
% \resizebox{\textwidth}{!}{
% \begin{minipage}{\columnwidth}
% \includegraphics[width=\columnwidth]{figure/llm_as_a_judge.png}
% \end{minipage}
% }
% \caption{Evaluation prompts for LLM-as-a-Judge.}
% \label{fig:llm_as_a_judge_prompt}
% \end{figure*}
% % ------------------------------------------

\section{Additional Experiments}
\label{app:additional_experiment}

% \subsection{Models}
% \label{app:model}
% In addition to the LLMs listed in Section~\ref{subsec:models}, we further validate our findings using models from a different family, specifically \lm{Llama3-8B} and \lm{Llama3-70B} \cite{llama3modelcard}. Detailed information about these two models is provided in Table~\ref{tab:used_model}.

\subsection{Knowledge Memorization Analysis}
\label{app:memorization}

\input{table/factual_recall}
Table~\ref{tab:lre_acc} illustrates knowledge recall accuracy results on the \data{LRE} dataset for \texttt{Qwen2.5-\{7B,14B\}} and \texttt{Llama3-8B} models across different quantization methods.

\paragraph{Knowledge Recall Accuracy.}

In Table~\ref{tab:per-relation-factual-recall-acc-7B}, and \ref{tab:per-relation-factual-recall-acc-14B}, we provide the per-relation factual recall accuracy on \lm{Qwen2.5-7B}, \lm{Qwen2.5-14B}, and \lm{Llama3-8B} models, respectively. As discussion in Section~\ref{subsec:knowlegde_results}, we observe that the accuracy drop is more severe in relations where the original performance has not yet saturated.

\paragraph{Neuron Attribution Analysis.}
% \begin{figure*}[t]
%     \centering
%     \begin{minipage}{0.95\textwidth}
%         \begin{minipage}{0.48\textwidth}
%             \centering
%         \includegraphics[width=\textwidth]{figure/relationship_comparison/landmark_on_continent_methods_comparison.pdf}
%         \subcaption{Contribution score comparison}
%         \end{minipage}%
%         \hfill%
%         \begin{minipage}{0.48\textwidth}
%             \centering
%         \includegraphics[width=\textwidth]{figure/neuron_distribution/landmark_on_continent_layer_distribution.pdf} 
%         \subcaption{Layer distribution}
%             \end{minipage}
%     \end{minipage}
%     \caption{Analysis on the top-300 neurons with highest contribution scores for the \textit{landmark on continent} relation under different quantization methods. (a) Average contribution scores of top 300 feed-forward neurons across different quantization methods, showing how each method affects neuron activation patterns. (b) Distribution of high-scoring neurons across model layers, showing the number of neurons that exceed the full-precision model's 300th neuron score threshold in each layer.}
%     \label{fig:landmark_continent_analysis}
% \end{figure*}

\begin{figure*}[t!]
    \centering
    % Use figure* to span both columns
    
    % Row 2: Person father
    \begin{minipage}{0.95\textwidth}
        \begin{minipage}{0.48\textwidth}
            \centering
            \includegraphics[width=\textwidth]{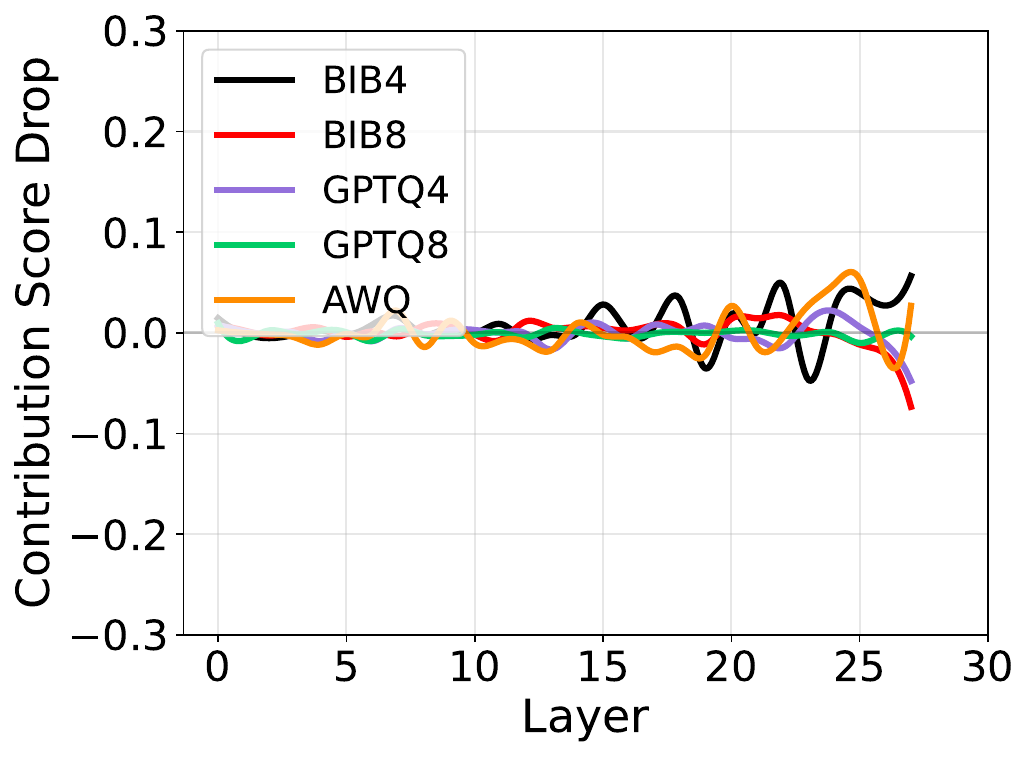}
            \subcaption{Attention: Person father}
        \end{minipage}%
        \hfill%
        \begin{minipage}{0.48\textwidth}
            \centering
            \includegraphics[width=\textwidth]{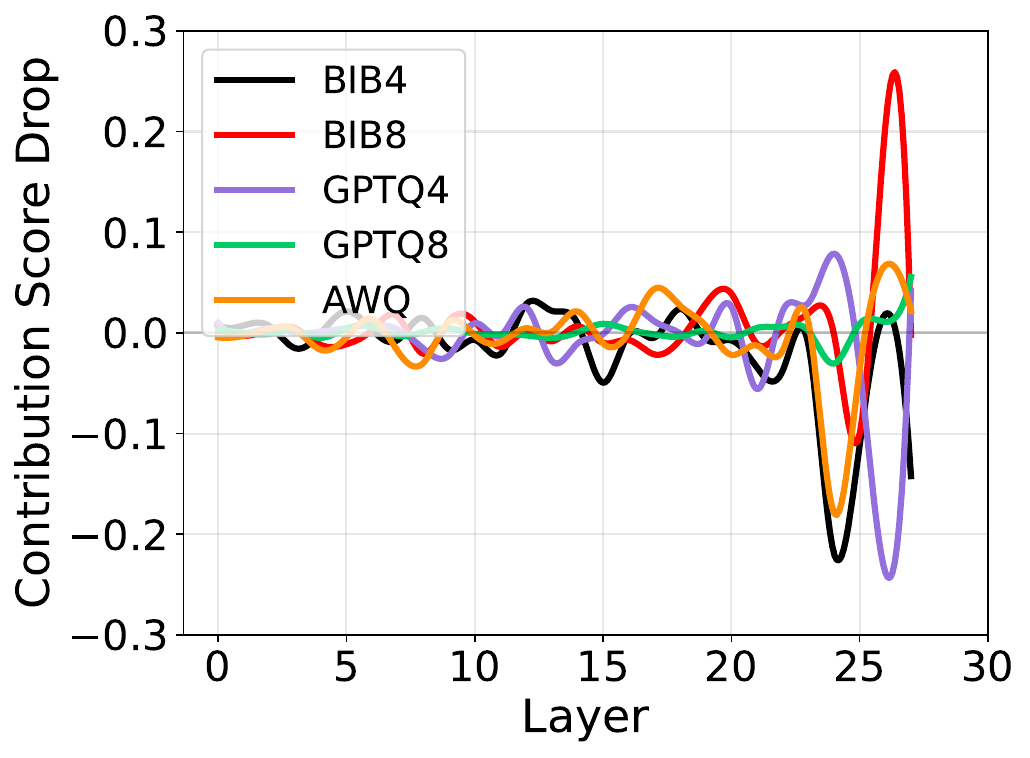}
            \subcaption{FFN: Person father}
        \end{minipage}
    \end{minipage}
    
    \vspace{0.4cm}
    
    % Row 3: Person mother
    \begin{minipage}{0.95\textwidth}
        \begin{minipage}{0.48\textwidth}
            \centering
            \includegraphics[width=\textwidth]{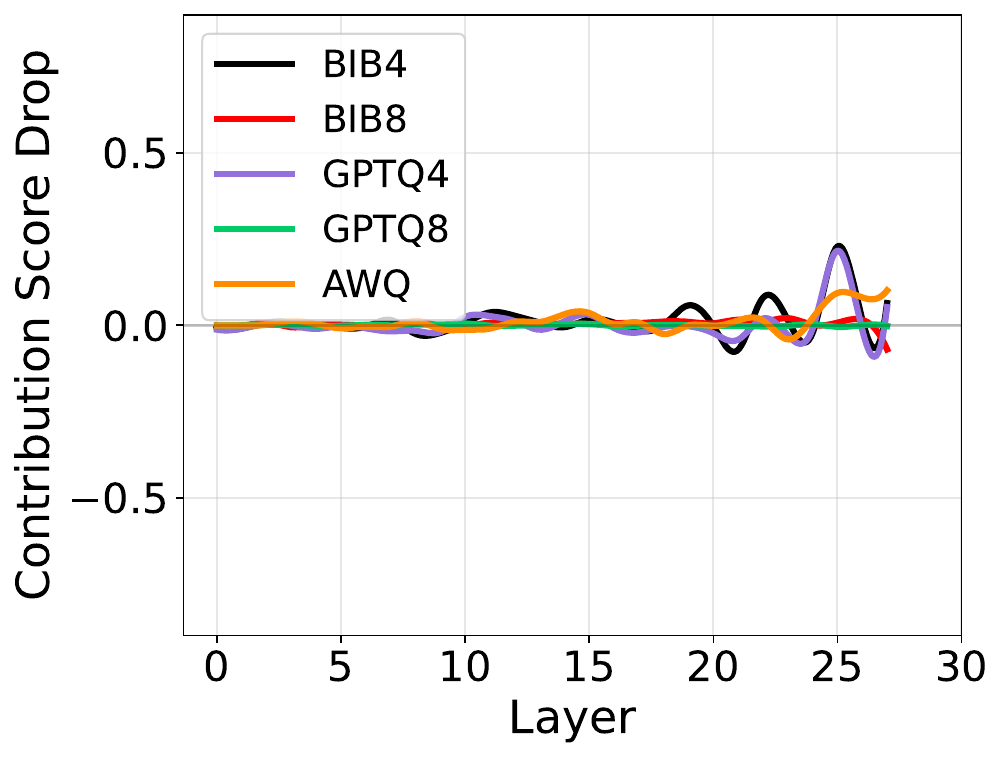}
            \subcaption{Attention: Person mother}
        \end{minipage}%
        \hfill%
        \begin{minipage}{0.48\textwidth}
            \centering
            \includegraphics[width=\textwidth]{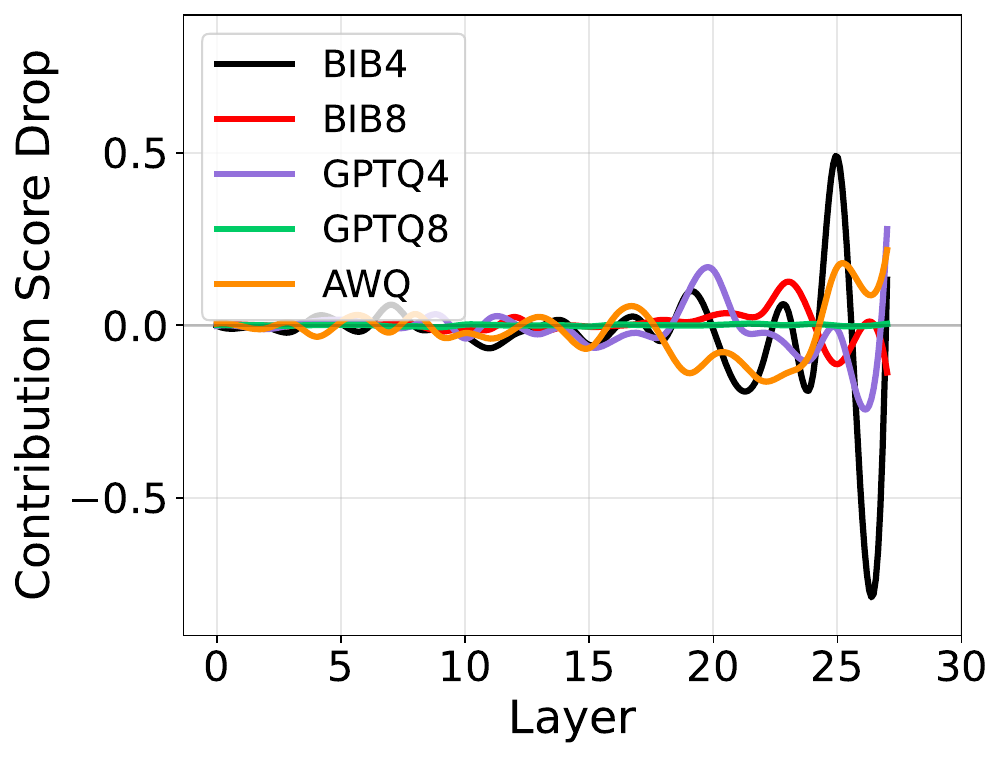}
            \subcaption{FFN: Person mother}
        \end{minipage}
    \end{minipage}
    
    \vspace{0.4cm}
    
    % Row 4: Person sport position
    \begin{minipage}{0.95\textwidth}
        \begin{minipage}{0.48\textwidth}
            \centering
            \includegraphics[width=\textwidth]{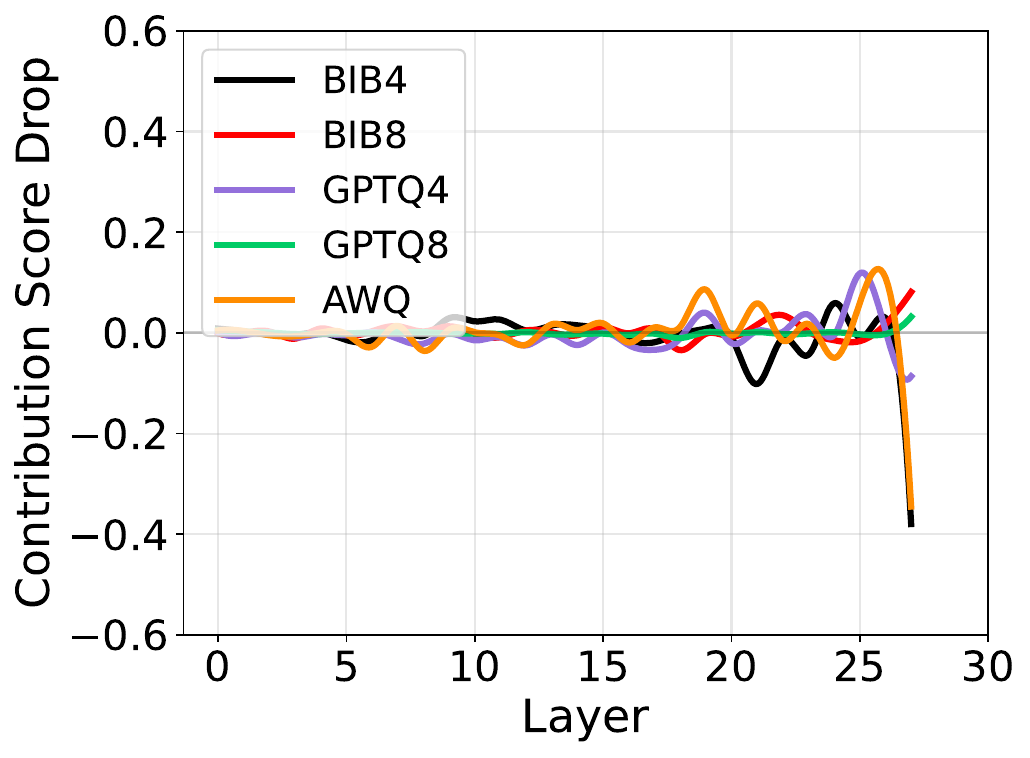}
            \subcaption{Attention: Person sport position}
        \end{minipage}%
        \hfill%
        \begin{minipage}{0.48\textwidth}
            \centering
            \includegraphics[width=\textwidth]{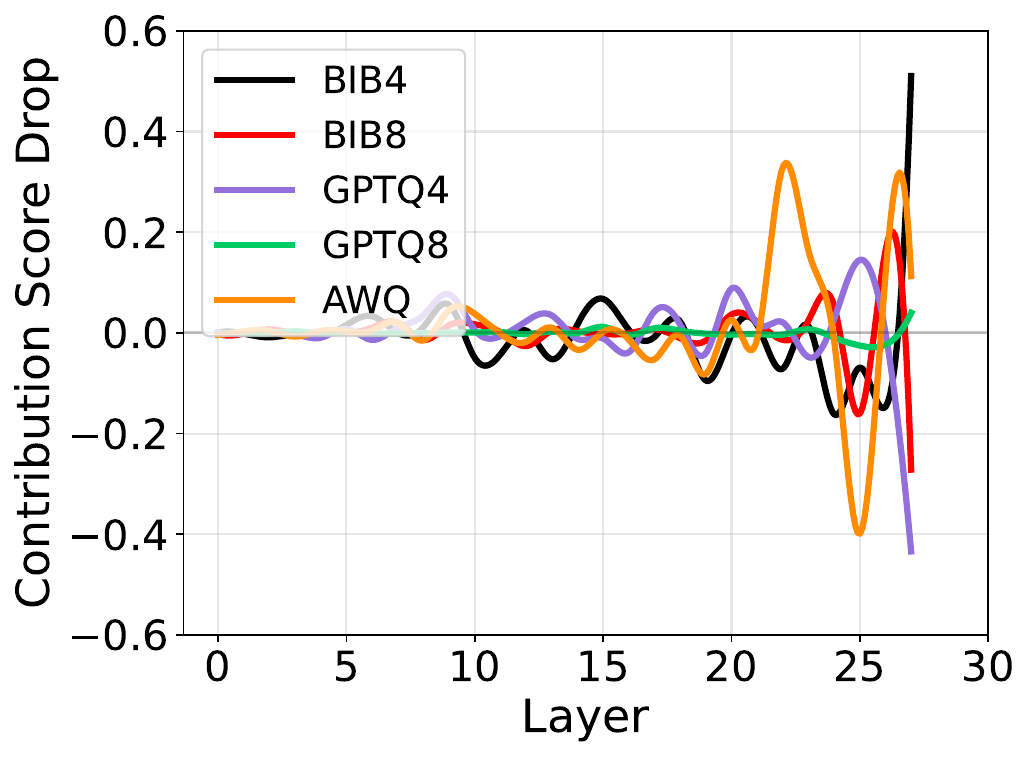}
            \subcaption{FFN: Person sport position}
        \end{minipage}
    \end{minipage}
    
    \caption{Contribution score drop across quantization methods on \lm{Qwen2.5-7B} for other relationship types, comparing attention sublayers (left) and feed-forward sublayers (right). Rows show different relationships: (a-b) person father, (c-d) person mother, and (e-f) person sport position. The plots reveal how different quantization methods affect knowledge representation across model layers.}
    \label{fig:layer_contribution_comparison_all}
    
\end{figure*}

\begin{figure*}[t!]
    \centering
    % Use figure* to span both columns
    
    % Row 2: Person father
    \begin{minipage}{0.95\textwidth}
        \begin{minipage}{0.48\textwidth}
            \centering
            \includegraphics[width=\textwidth]{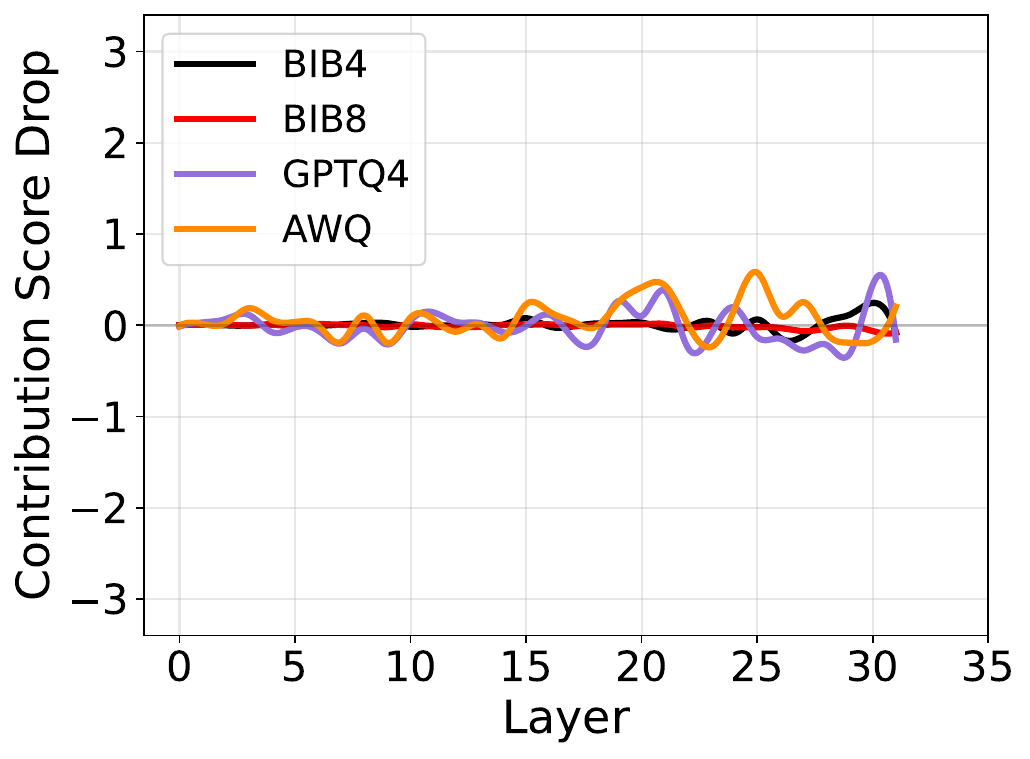}
            \subcaption{Attention: Person father}
        \end{minipage}%
        \hfill%
        \begin{minipage}{0.48\textwidth}
            \centering
            \includegraphics[width=\textwidth]{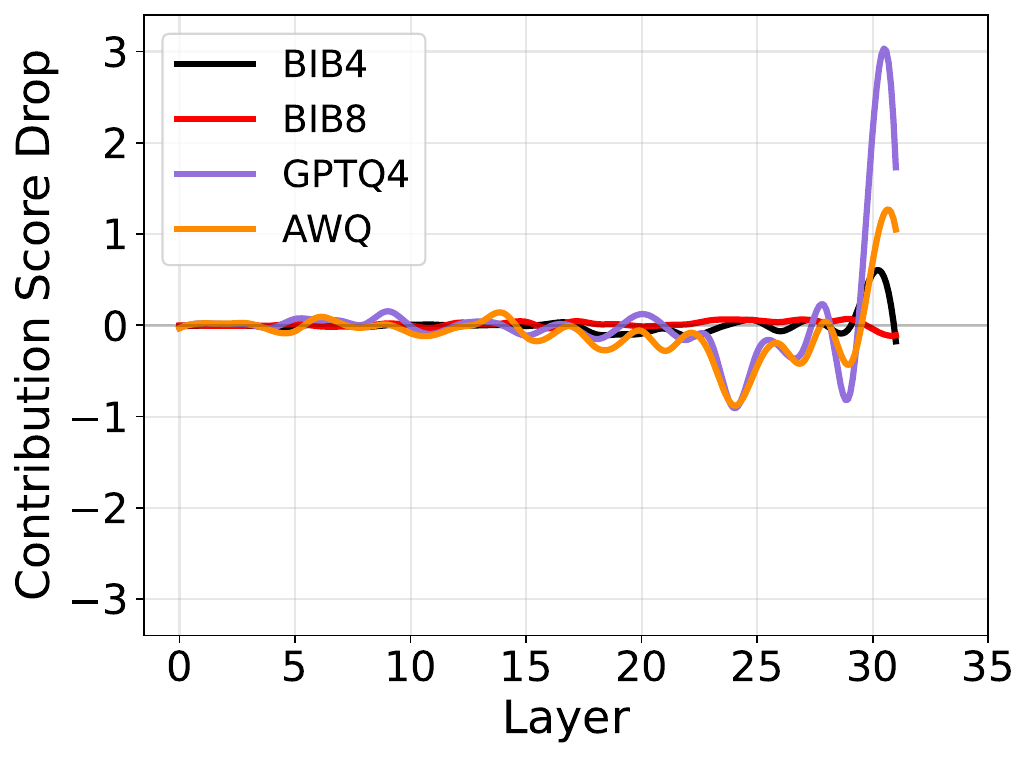}
            \subcaption{FFN: Person father}
        \end{minipage}
    \end{minipage}
    
    \vspace{0.4cm}
    
    % Row 3: Person mother
    \begin{minipage}{0.95\textwidth}
        \begin{minipage}{0.48\textwidth}
            \centering
            \includegraphics[width=\textwidth]{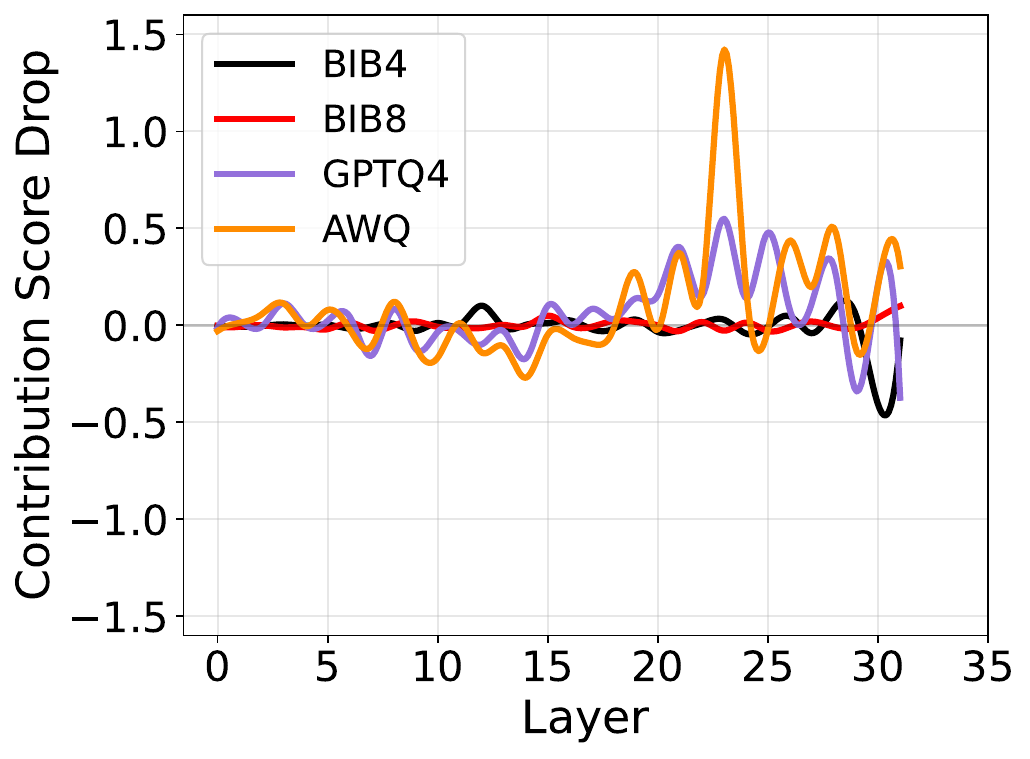}
            \subcaption{Attention: Person mother}
        \end{minipage}%
        \hfill%
        \begin{minipage}{0.48\textwidth}
            \centering
            \includegraphics[width=\textwidth]{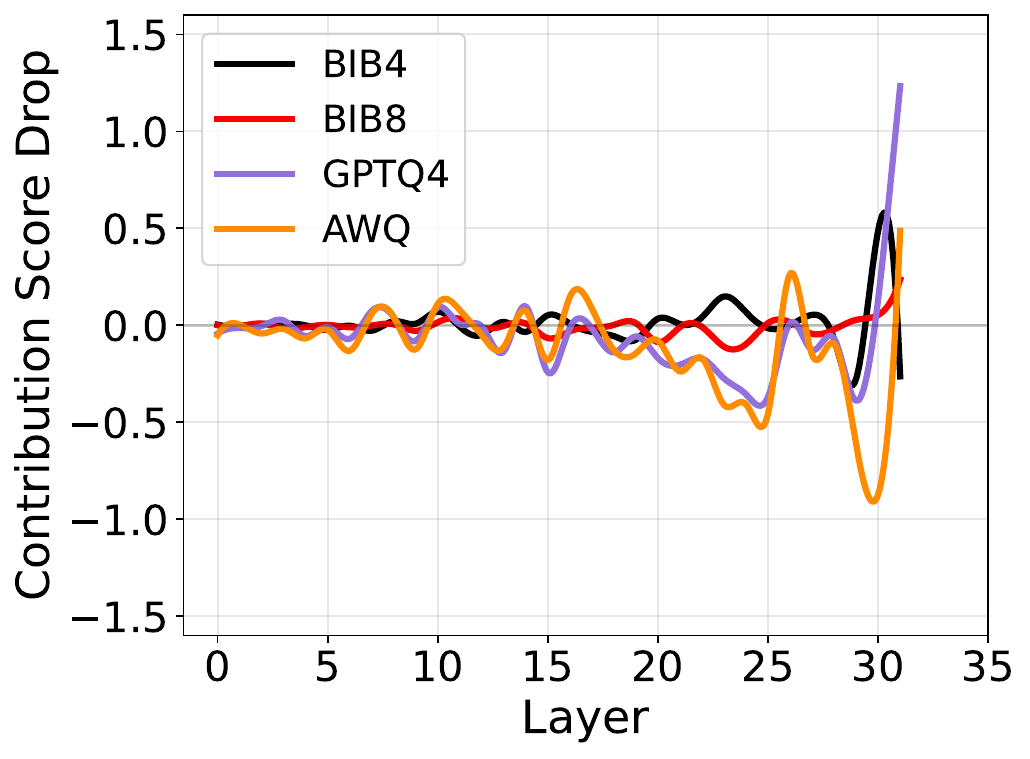}
            \subcaption{FFN: Person mother}
        \end{minipage}
    \end{minipage}
    
    \vspace{0.4cm}
    
    % Row 4: Person sport position
    \begin{minipage}{0.95\textwidth}
        \begin{minipage}{0.48\textwidth}
            \centering
            \includegraphics[width=\textwidth]{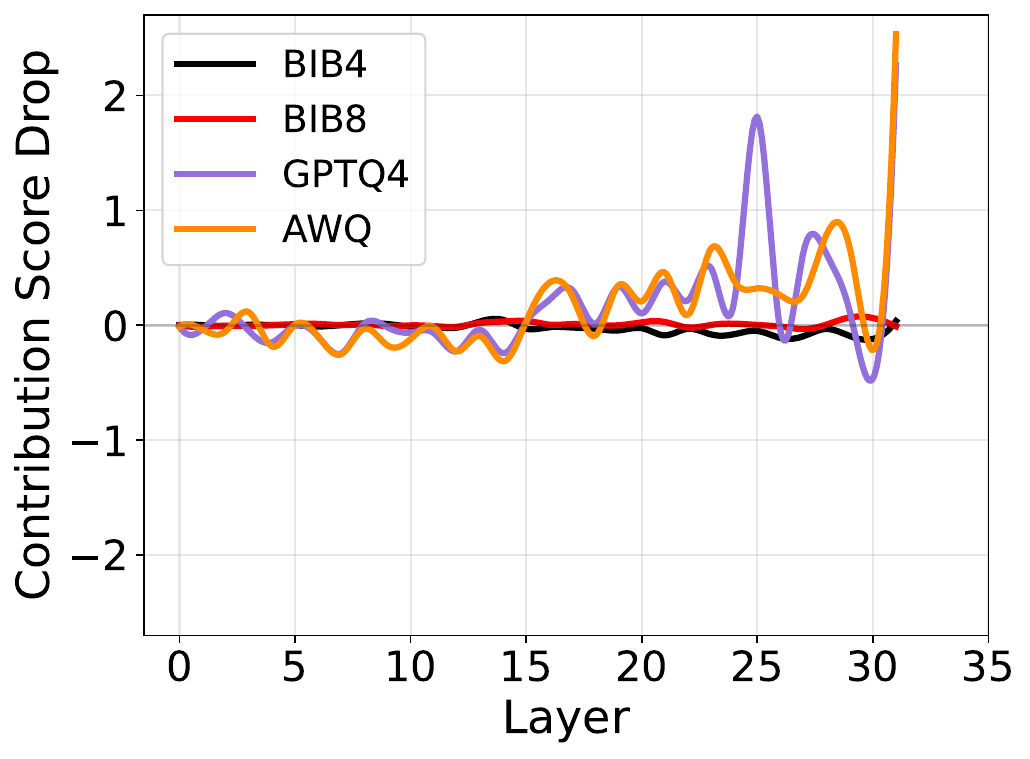}
            \subcaption{Attention: Person sport position}
        \end{minipage}%
        \hfill%
        \begin{minipage}{0.48\textwidth}
            \centering
            \includegraphics[width=\textwidth]{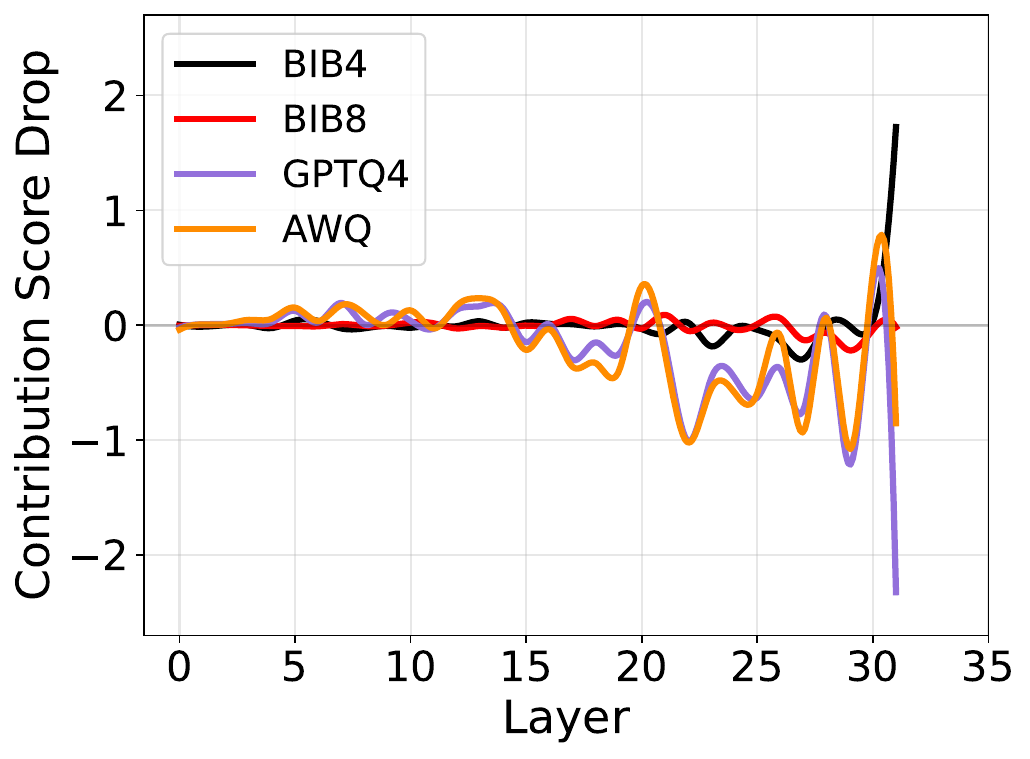}
            \subcaption{FFN: Person sport position}
        \end{minipage}
    \end{minipage}
    
    \caption{Contribution score drop across quantization methods on \lm{Llama3-8B} for other relationship types, comparing attention sublayers (left) and feed-forward sublayers (right). Rows show different relationships: (a-b) person father, (c-d) person mother, and (e-f) person sport position. The plots reveal how different quantization methods affect knowledge representation across model layers.}
    \label{fig:layer_contribution_comparison_all_llama}
    
\end{figure*}

\begin{figure*}[t]
    \centering
    
    % Row 2: Person father
    \begin{tabular}{cc}
        \includegraphics[width=0.48\textwidth]{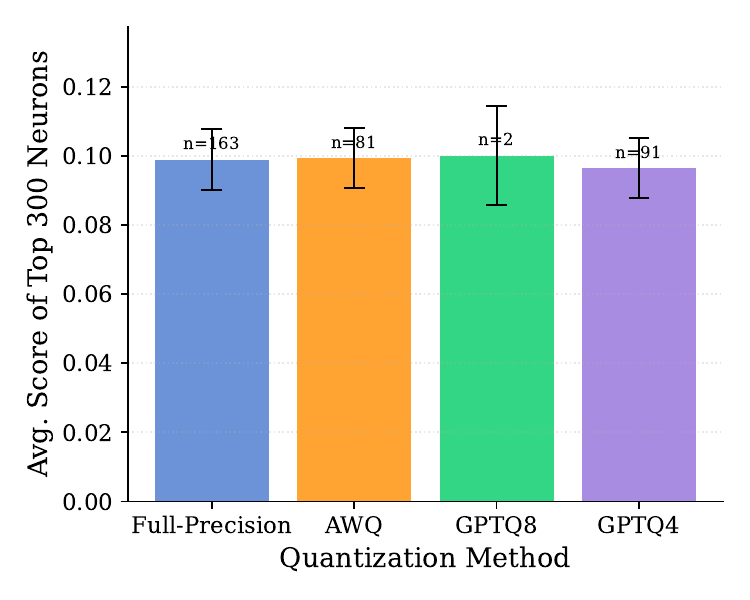} &
        \includegraphics[width=0.48\textwidth]{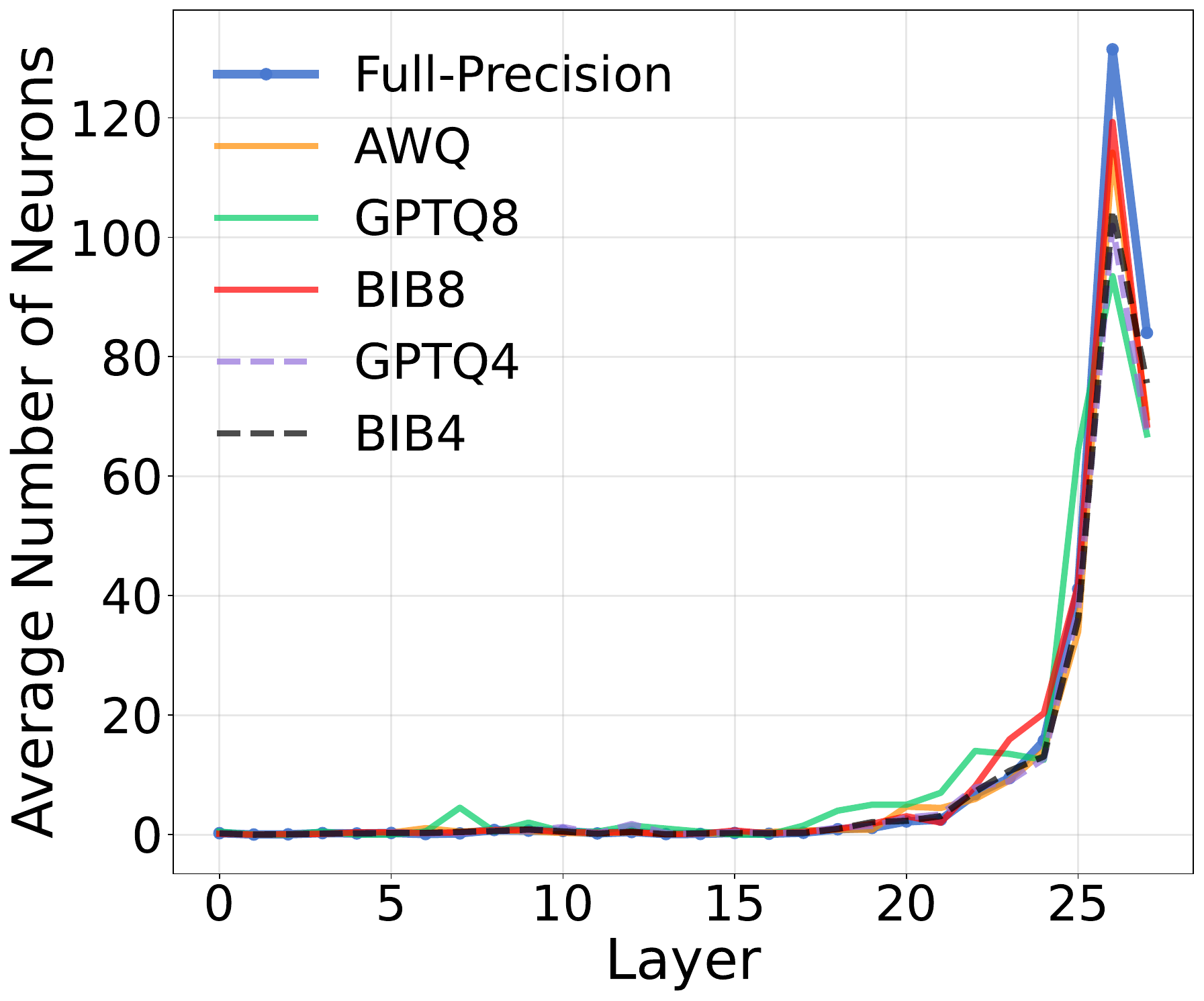} \\
        \footnotesize (a) Person father: Contribution scores & \footnotesize (b) Person father: Layer distribution \\
    \end{tabular}
    
    \vspace{0.3cm}
    
    % Row 3: Person mother
    \begin{tabular}{cc}
        \includegraphics[width=0.48\textwidth]{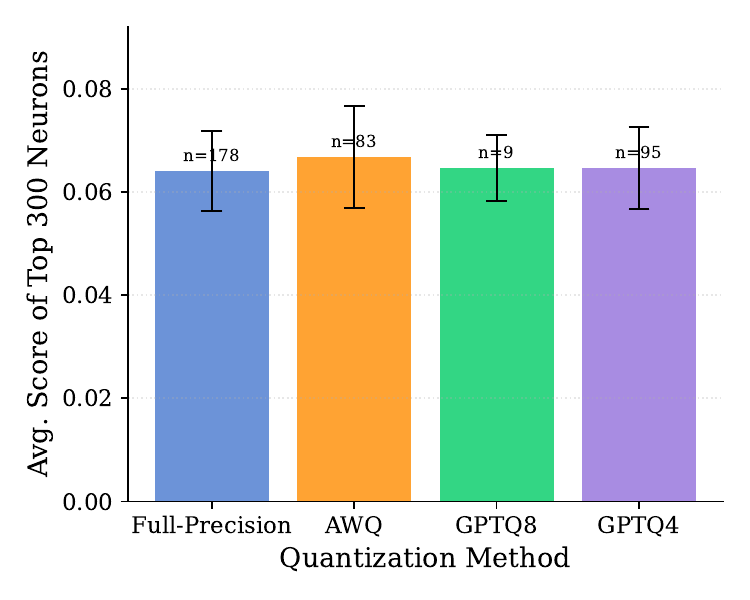} &
        \includegraphics[width=0.48\textwidth]{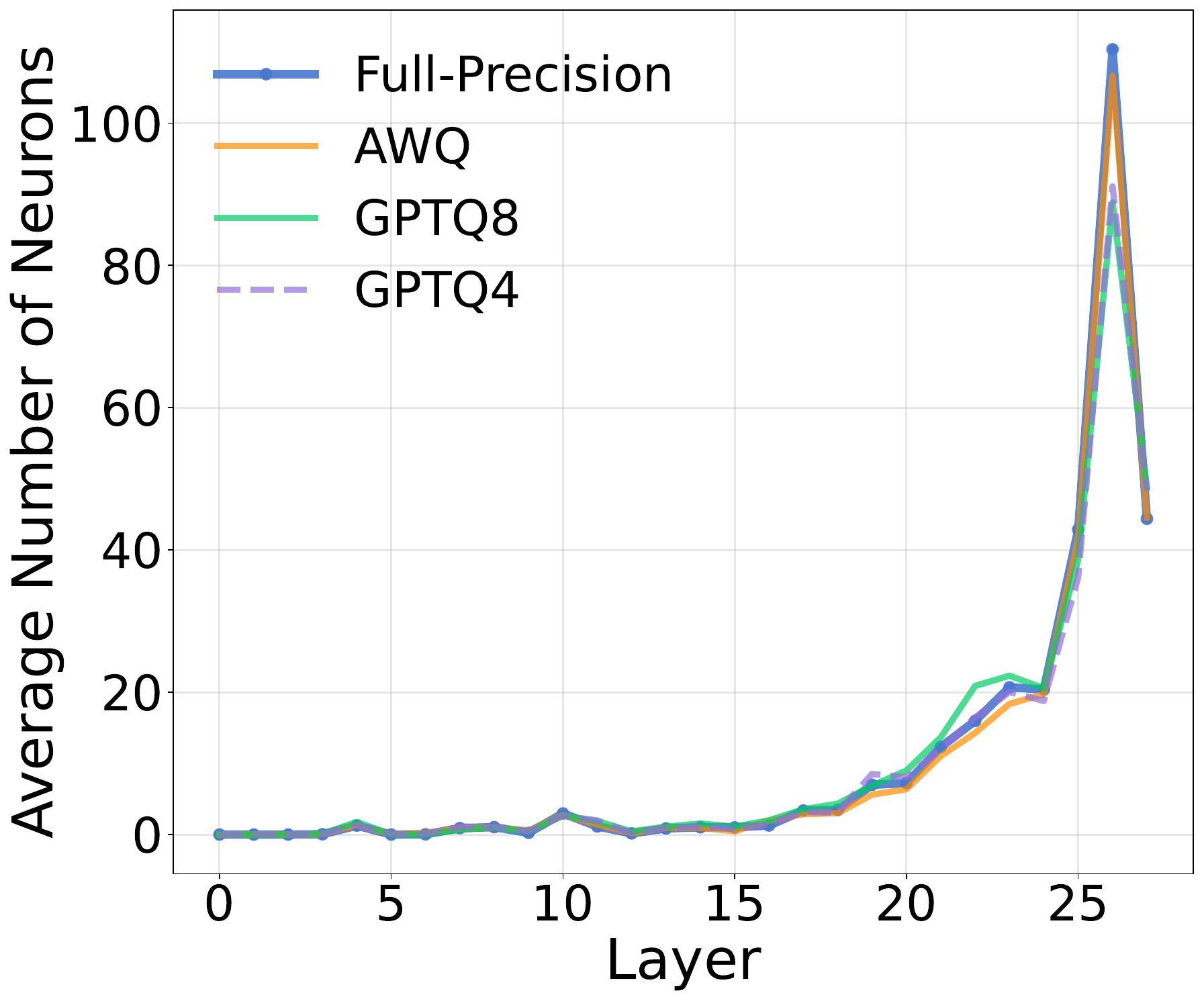} \\
        \footnotesize (c) Person mother: Contribution scores & \footnotesize (d) Person mother: Layer distribution \\
    \end{tabular}
    
    \vspace{0.3cm}
    
    % Row 4: Person sport position
    \begin{tabular}{cc}
        \includegraphics[width=0.48\textwidth]{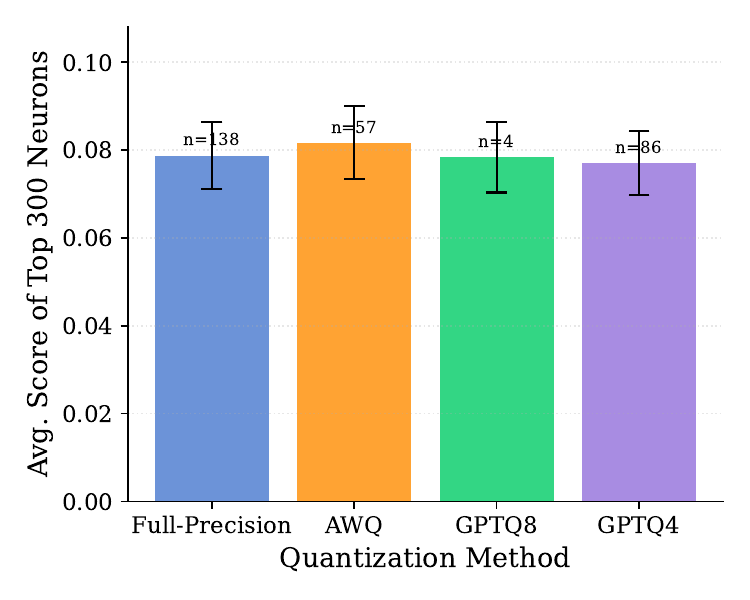} &
        \includegraphics[width=0.48\textwidth]{figure/neuron_distribution/person_sport_position_layer_distribution.pdf} \\
        \footnotesize (e) Person sport position: Contribution scores & \footnotesize (f) Person sport position: Layer distribution \\
    \end{tabular}
    
    \caption{Analysis of other relationship types under different quantization methods applied to \lm{Qwen2.5-7B}. Left column: Average contribution scores of top 300 feed-forward neurons across quantization methods. Right column: Distribution of high-scoring neurons across model layers. Each row represents a different relationship: (a-b) person father, (c-d) person mother, and (e-f) person sport position. These visualizations reveal both the magnitude of contribution score changes and their distribution across the model architecture when applying different quantization techniques to various types of factual knowledge.}
    \label{fig:all_relationships_analysis}
\end{figure*}

\begin{figure*}[t]
    \centering
    
    % Row 2: Person father
    \begin{tabular}{cc}
        \includegraphics[width=0.48\textwidth]{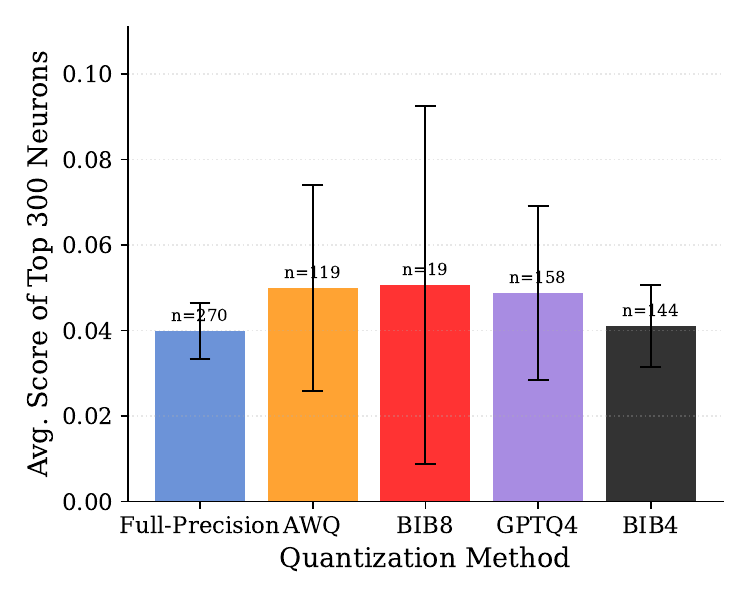} &
        \includegraphics[width=0.48\textwidth]{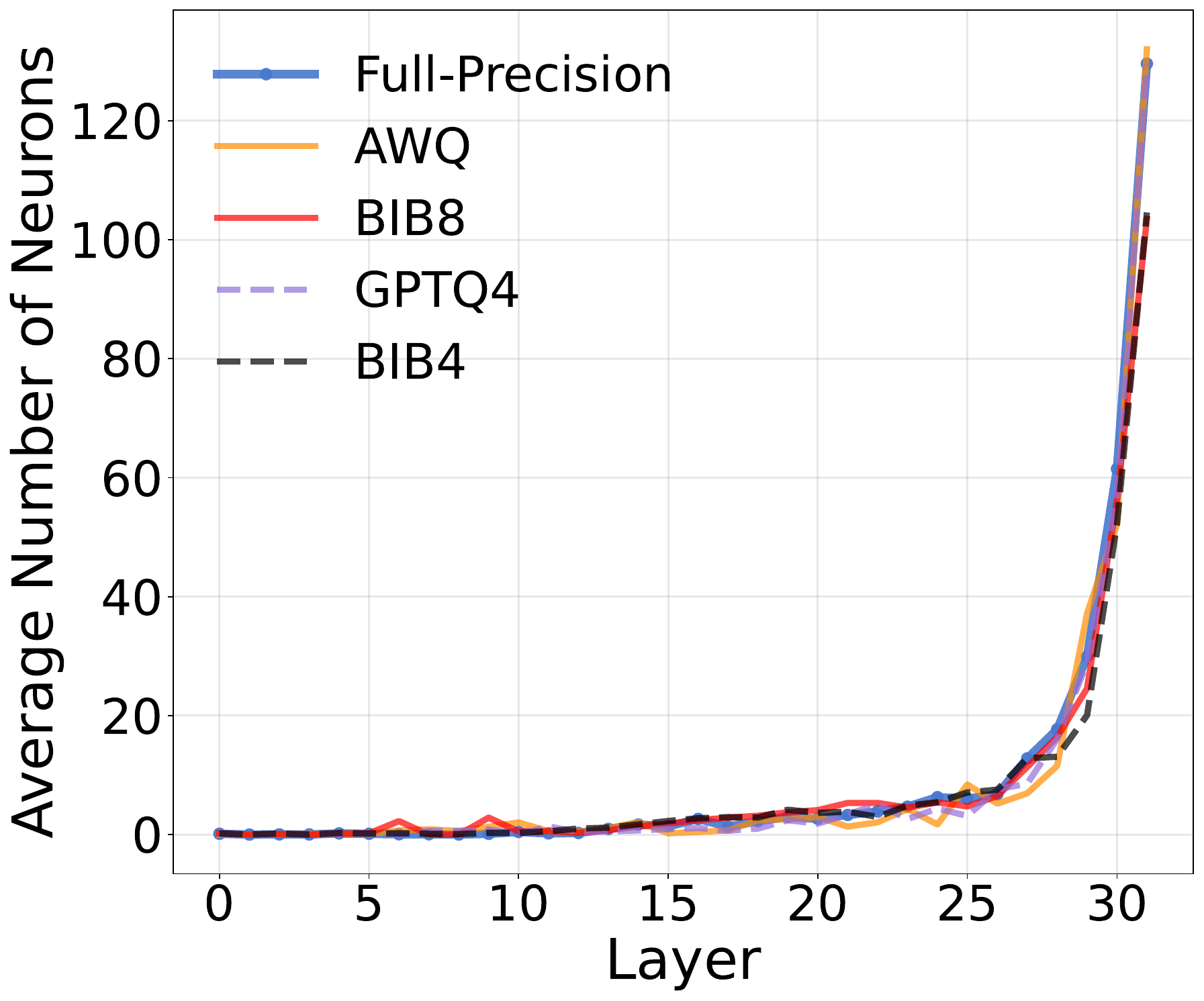} \\
        \footnotesize (a) Person father: Contribution scores & \footnotesize (b) Person father: Layer distribution \\
    \end{tabular}
    
    \vspace{0.3cm}
    
    % Row 3: Person mother
    \begin{tabular}{cc}
        \includegraphics[width=0.48\textwidth]{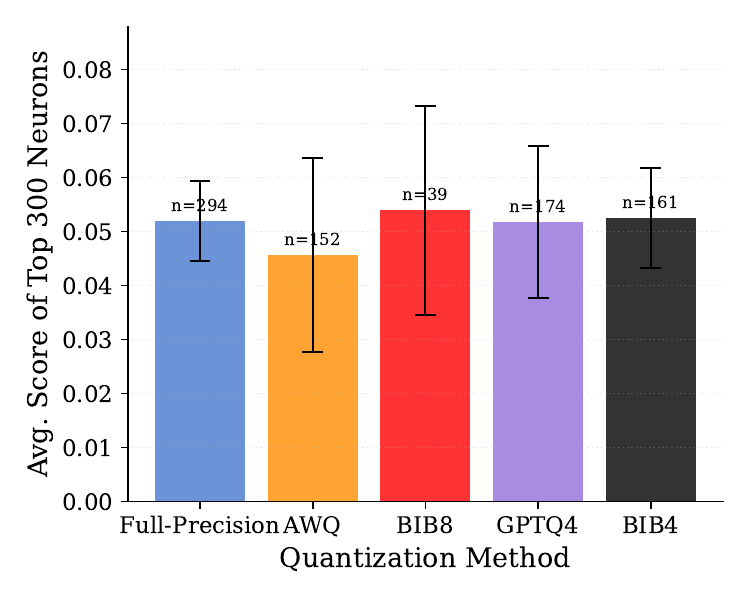} &
        \includegraphics[width=0.48\textwidth]{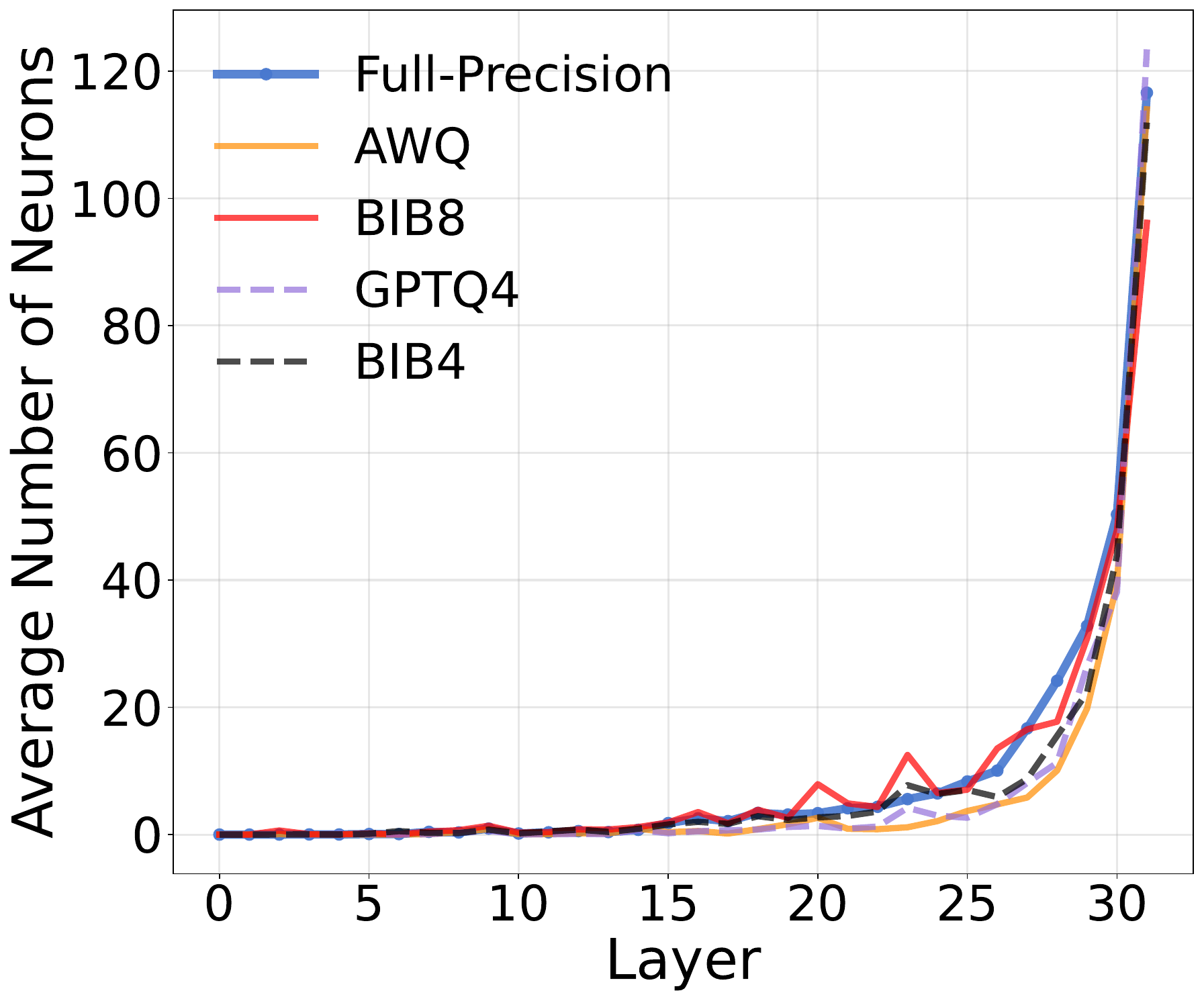} \\
        \footnotesize (c) Person mother: Contribution scores & \footnotesize (d) Person mother: Layer distribution \\
    \end{tabular}
    
    \vspace{0.3cm}
    
    % Row 4: Person sport position
    \begin{tabular}{cc}
        \includegraphics[width=0.48\textwidth]{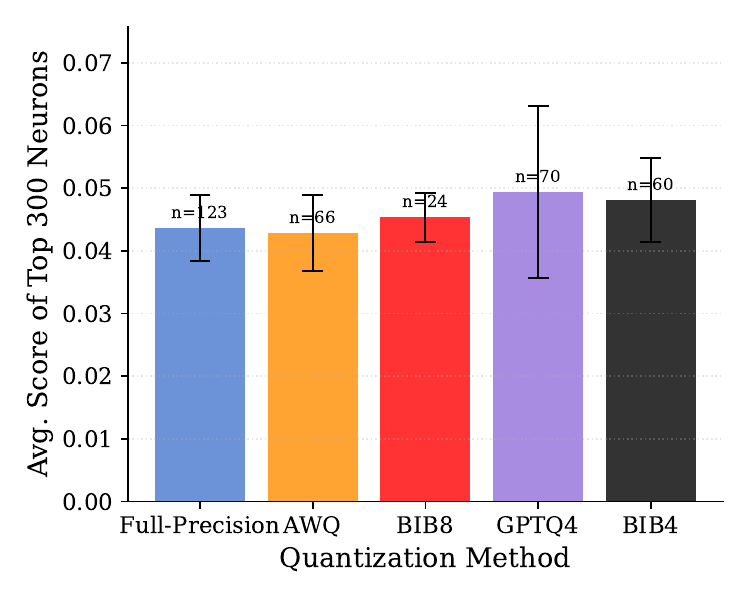} &
        \includegraphics[width=0.48\textwidth]{figure/neuron_distribution/llama/person_sport_position_layer_distribution.pdf} \\
        \footnotesize (e) Person sport position: Contribution scores & \footnotesize (f) Person sport position: Layer distribution \\
    \end{tabular}
    
    \caption{Analysis of other relationship types under different quantization methods applied to \lm{Llama3-8B}. Left column: Average contribution scores of top 300 feed-forward neurons across quantization methods. Right column: Distribution of high-scoring neurons across model layers. Each row represents a different relationship: (a-b) person father, (c-d) person mother, and (e-f) person sport position. These visualizations reveal both the magnitude of contribution score changes and their distribution across the model architecture when applying different quantization techniques to various types of factual knowledge.}
    \label{fig:all_relationships_analysis_llama}
\end{figure*}

% For two-column IEEE format - spans both columns and centers content
Here we present the neuron-level knowledge attribution results on relations \textit{person father}, \textit{person mother}, \textit{person sport position}, complementing our results in Section~\ref{subsec:knowlegde_results}. Analysis on layer-wise neuron contribution scores are shown in Figure~\ref{fig:layer_contribution_comparison_all} and Figure~\ref{fig:layer_contribution_comparison_all_llama}, the analysis on top-300 neurons are given in Figure~\ref{fig:all_relationships_analysis} and Figure~\ref{fig:all_relationships_analysis_llama}.

\begin{figure*}[t]
    \centering
    \begin{minipage}{\textwidth}
        \begin{minipage}{0.49\textwidth}
            \centering
        \includegraphics[width=\textwidth]{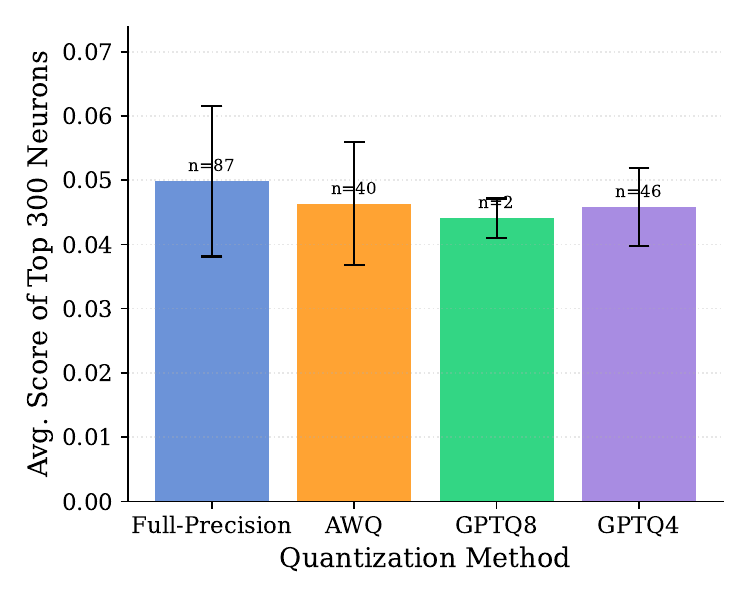}
        \subcaption{Contribution score comparison}
        \end{minipage}%
        \hfill%
        \begin{minipage}{0.49\textwidth}
            \centering
        \includegraphics[width=\textwidth]{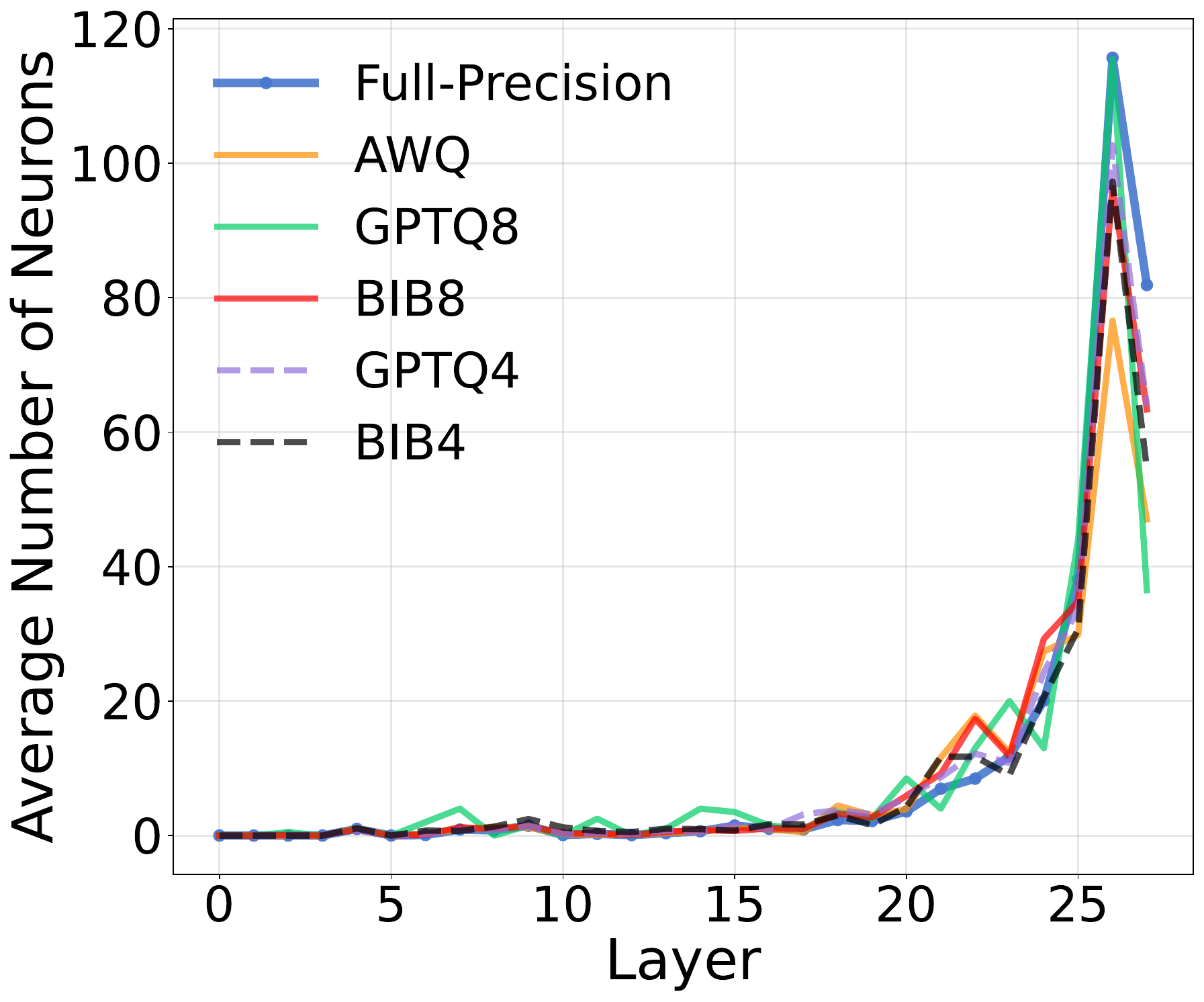} 
        \subcaption{Layer distribution}
            \end{minipage}
    \end{minipage}
    \caption{Analysis of the top-300 neurons with highest contribution scores for the \textit{landmark on continent} relation under different quantization methods applied to \lm{Qwen2.5-7B}. (a) Average contribution scores of top 300 feed-forward neurons across different quantization methods, showing how each method affects neuron activation patterns. (b) Distribution of high-scoring neurons across layers, showing the number of neurons exceeding the full-precision model's 300th neuron score threshold in each layer.}
    \label{fig:landmark_continent_analysis}
\end{figure*}

\begin{figure*}[t]
    \centering
    \begin{minipage}{\textwidth}
        \begin{minipage}{0.49\textwidth}
            \centering
        \includegraphics[width=\textwidth]{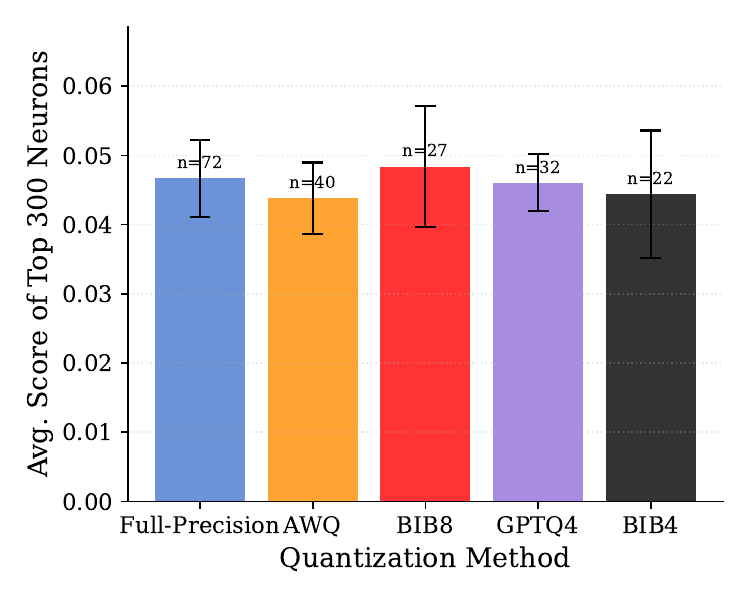}
        \subcaption{Contribution score comparison}
        \end{minipage}%
        \hfill%
        \begin{minipage}{0.49\textwidth}
            \centering
        \includegraphics[width=\textwidth]{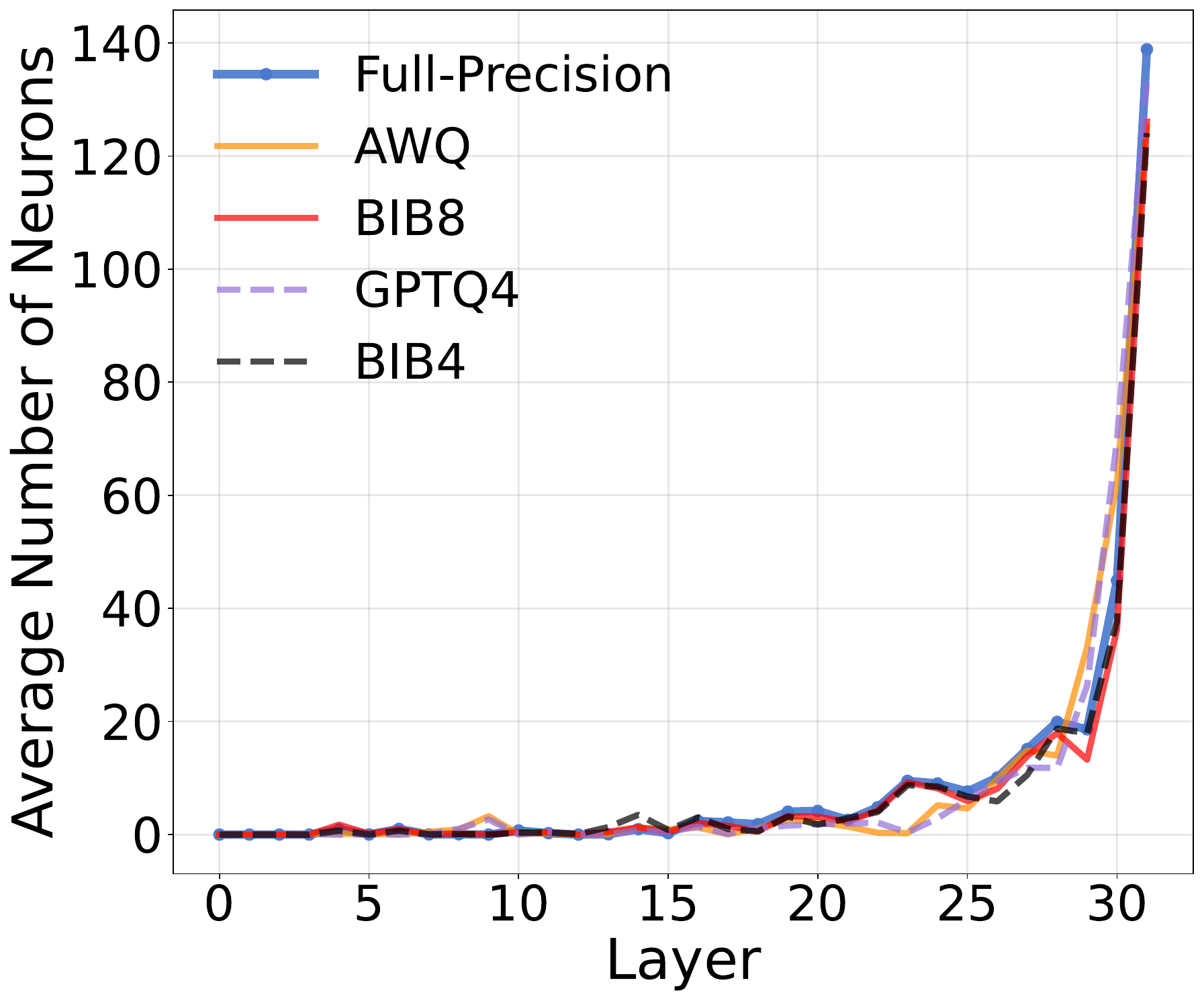} 
        \subcaption{Layer distribution}
            \end{minipage}
    \end{minipage}
    \caption{Analysis of the top-300 neurons with highest contribution scores for the \textit{landmark on continent} relation under different quantization methods applied to \lm{Llama3-8B}. (a) Average contribution scores of top 300 feed-forward neurons across different quantization methods, showing how each method affects neuron activation patterns. (b) Distribution of high-scoring neurons across layers, showing the number of neurons exceeding the full-precision model's 300th neuron score threshold in each layer.}
    \label{fig:landmark_continent_analysis_llama}
\end{figure*}

\subsection{Latent Multi-hop Reasoning}
\begin{figure*}[t!]
  \centering

  \begin{subfigure}{\textwidth}
    \centering
  \begin{subfigure}{\textwidth}
    \centering
    \centering
\includegraphics[width=\linewidth]{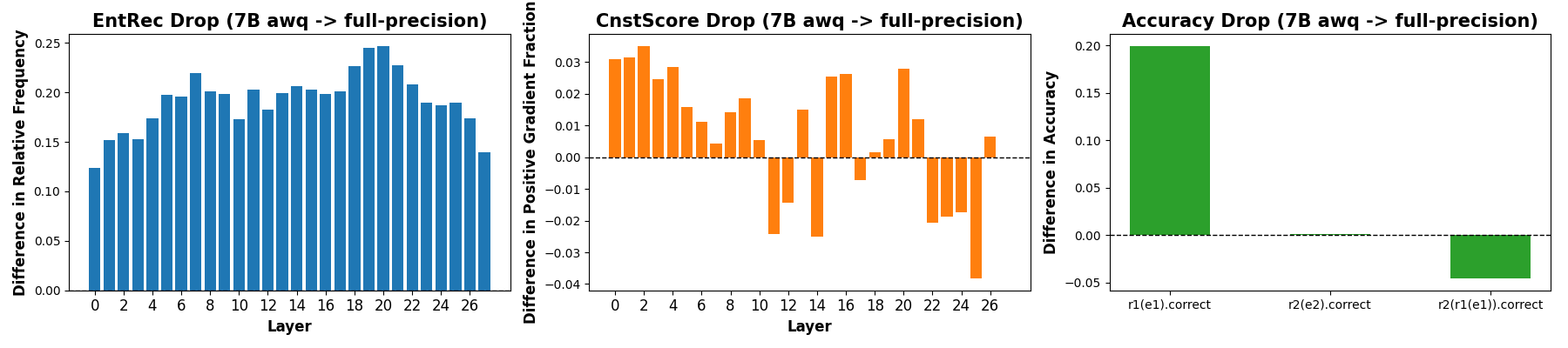}
\caption{\texttt{AWQ}}
  \end{subfigure}
  
  \hfill

\includegraphics[width=\linewidth]{figure/lmhr/7b_gptq8.png}
\caption{\texttt{GPTQ8}}
  \end{subfigure}
  
  \hfill

  \begin{subfigure}{\textwidth}
    \centering
    \centering
\includegraphics[width=\linewidth]{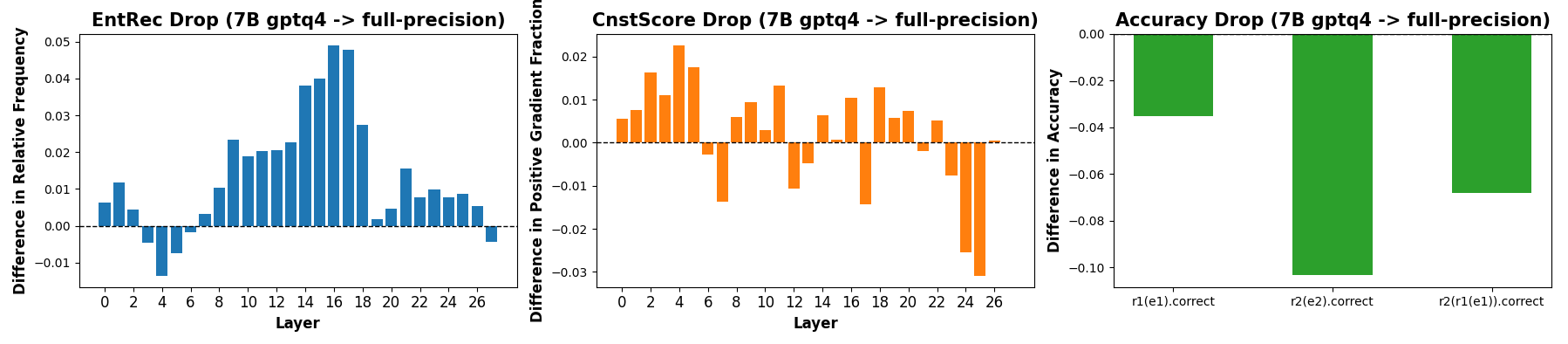}
\caption{\texttt{GPTQ4}}
  \end{subfigure}
  
  \hfill

    \begin{subfigure}{\textwidth}
    \centering
    \centering
\includegraphics[width=\linewidth]{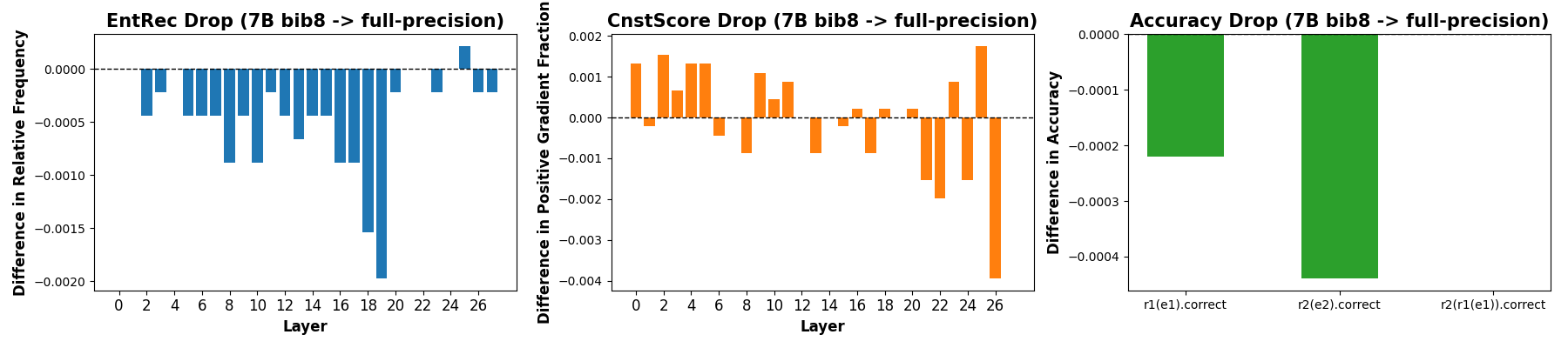}
\caption{\texttt{bib8}}
  \end{subfigure}
  
  \hfill

    \begin{subfigure}{\textwidth}
    \centering
    \centering
\includegraphics[width=\linewidth]{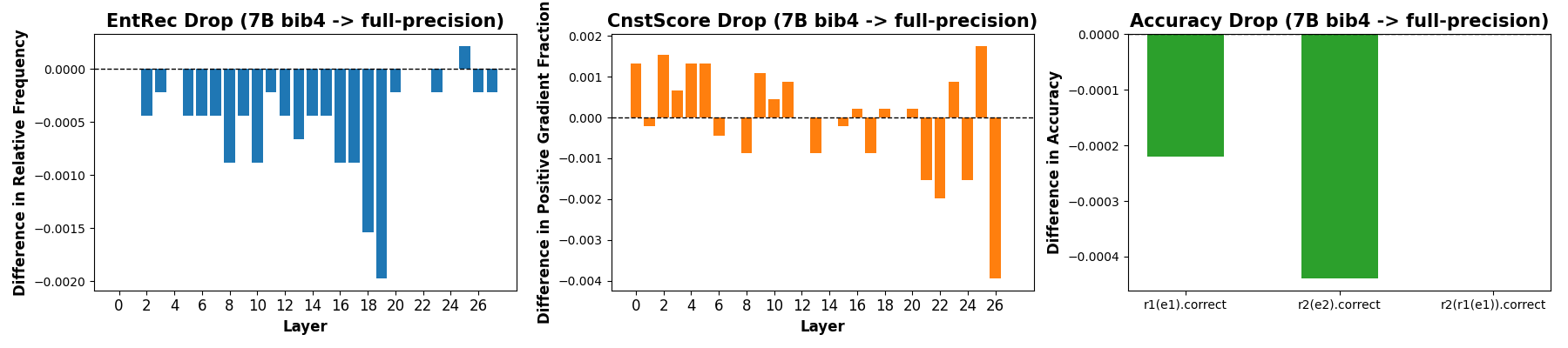}
\caption{\texttt{bib4}}
\label{fig:7b_bib4}

  \end{subfigure}
  
  \hfill

    \caption{Difference in the \textit{entity recall score} (\sys{EntRec}), \textit{consistency score} (\sys{CnstScore}), and \textit{accuracy} between the \texttt{AWQ}, \texttt{GPTQ8}, \texttt{GPTQ4}, \texttt{bib8}, \texttt{bib4} quantized and full-precision models of \lm{Qwen2.5-7B}, evaluated across all layers.}
  \label{fig:diff_7b}
\end{figure*}

\begin{figure*}[t!]
  \centering

  \begin{subfigure}{\textwidth}
    \centering
  \begin{subfigure}{\textwidth}
    \centering
    \centering
\includegraphics[width=\linewidth]{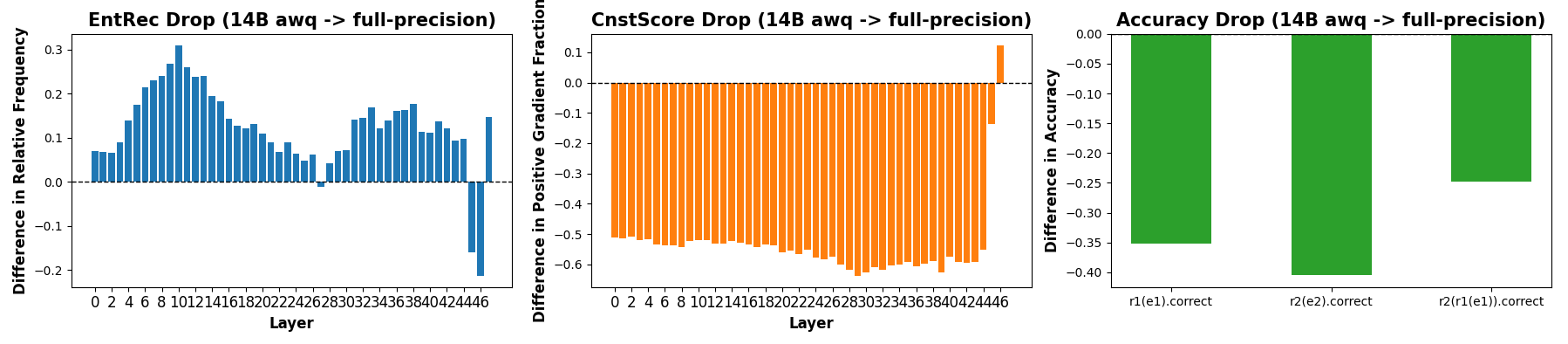}
\caption{\texttt{AWQ}}
  \end{subfigure}
  
  \hfill

\includegraphics[width=\linewidth]{figure/lmhr/14b_gptq8.png}
\caption{\texttt{GPTQ8}}
  \end{subfigure}
  
  \hfill

  \begin{subfigure}{\textwidth}
    \centering
    \centering
\includegraphics[width=\linewidth]{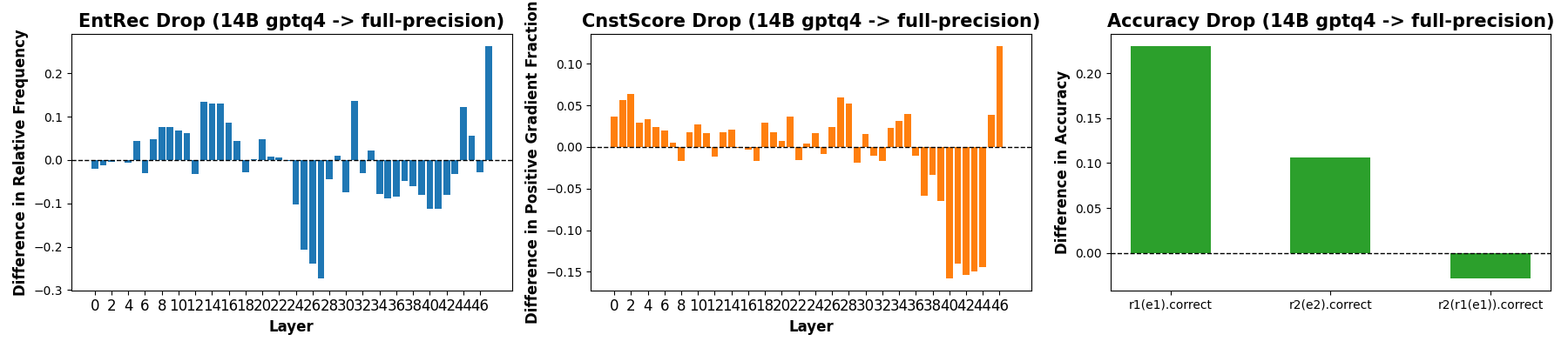}
\caption{\texttt{GPTQ4}}
  \end{subfigure}
  
  \hfill

    \begin{subfigure}{\textwidth}
    \centering
    \centering
\includegraphics[width=\linewidth]{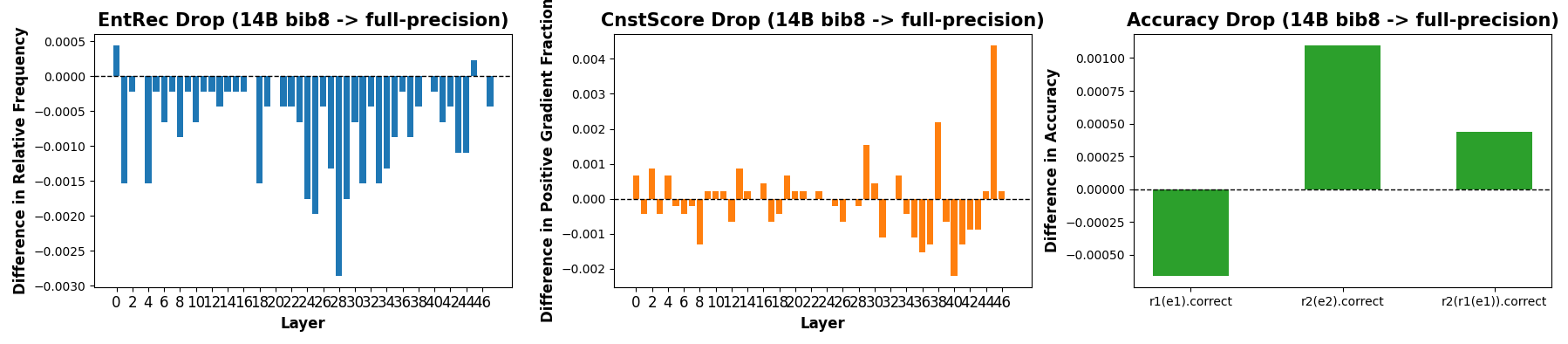}
\caption{\texttt{bib8}}
  \end{subfigure}
  
  \hfill

    \begin{subfigure}{\textwidth}
    \centering
    \centering
\includegraphics[width=\linewidth]{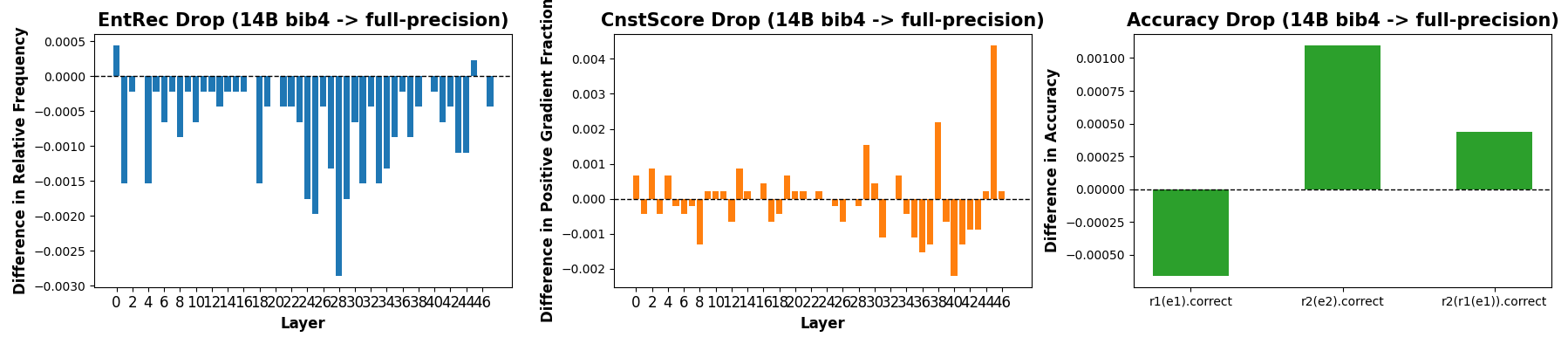}
\caption{\texttt{bib4}}
\label{fig:14b_bib4}
  \end{subfigure}
  
  \hfill

    \caption{Difference in the \textit{entity recall score} (\sys{EntRec}), \textit{consistency score} (\sys{CnstScore}), and \textit{accuracy} between the \texttt{AWQ}, \texttt{GPTQ8}, \texttt{GPTQ4}, \texttt{bib8}, \texttt{bib4} quantized and full-precision models of \lm{Qwen2.5-14B}, evaluated across all layers.}
  \label{fig:diff_14b}
\end{figure*}

\begin{figure*}[t!]
  \centering

  \begin{subfigure}{\textwidth}
    \centering
    \centering
\includegraphics[width=\linewidth]{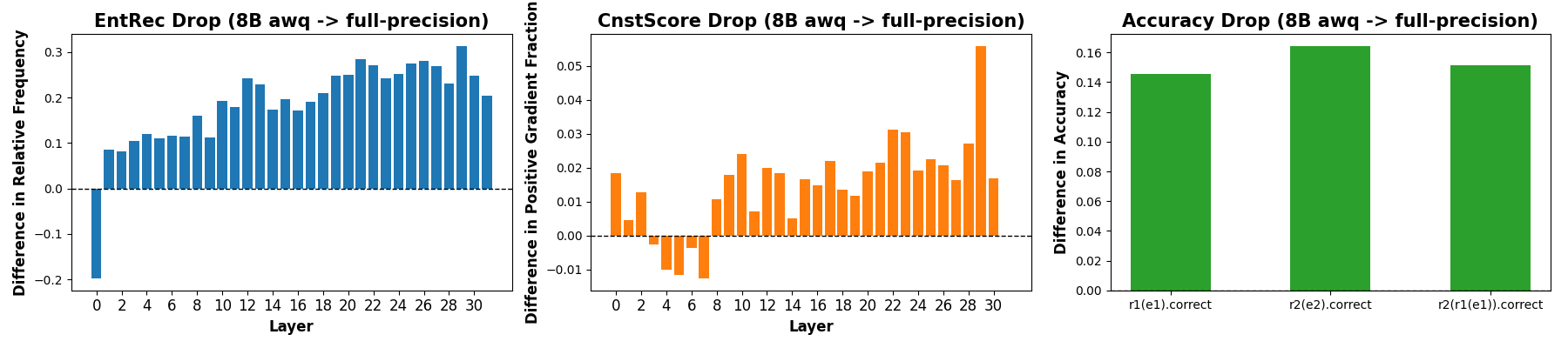}
\caption{\texttt{AWQ}}
  \end{subfigure}
  
  \hfill

  \begin{subfigure}{\textwidth}
    \centering
    \centering
\includegraphics[width=\linewidth]{figure/lmhr/8b_gptq.png}
\caption{\texttt{GPTQ}}
  \end{subfigure}
  
  \hfill

    \begin{subfigure}{\textwidth}
    \centering
    \centering
\includegraphics[width=\linewidth]{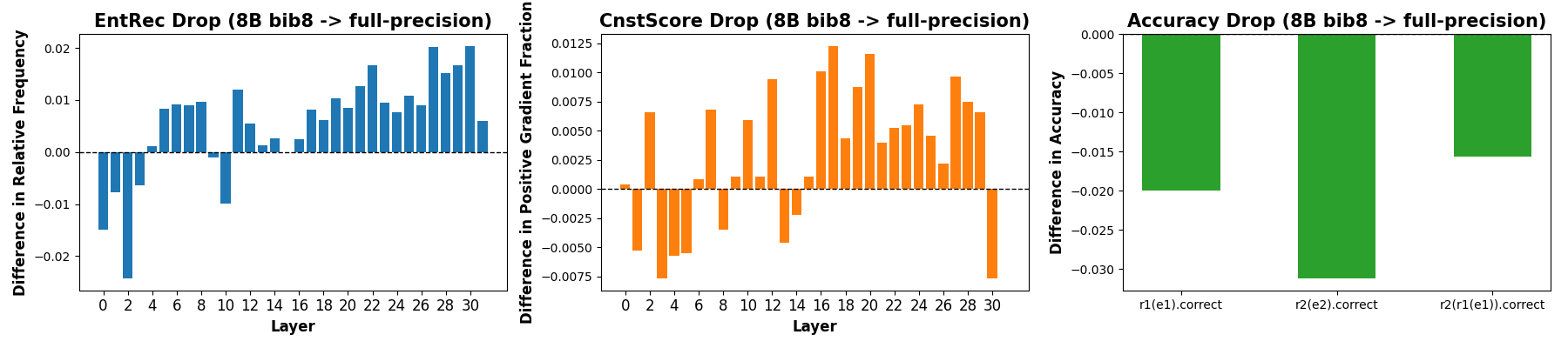}
\caption{\texttt{bib8}}
  \end{subfigure}
  
  \hfill

    \begin{subfigure}{\textwidth}
    \centering
    \centering
\includegraphics[width=\linewidth]{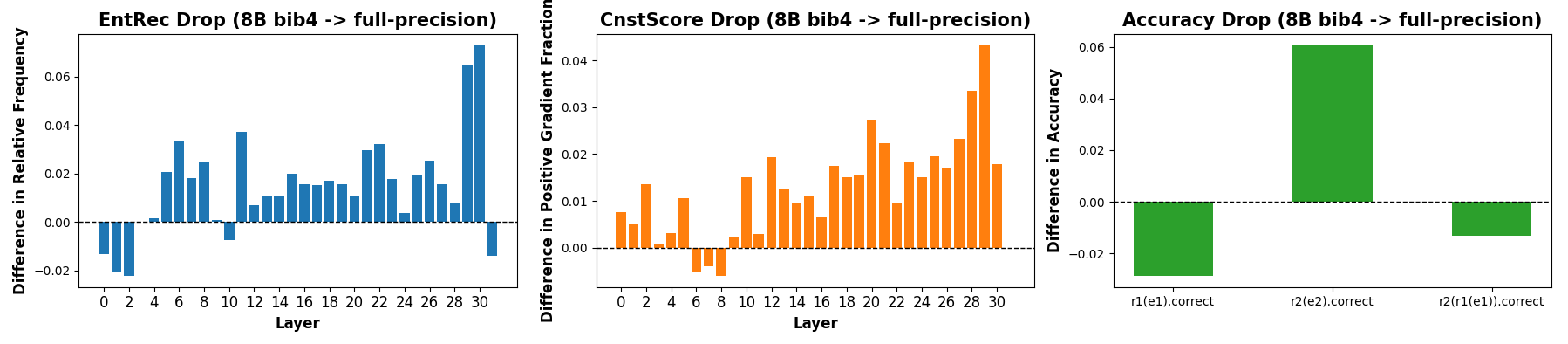}
\caption{\texttt{bib4}}
  \end{subfigure}
  
  \hfill

    \caption{Difference in the \textit{entity recall score} (\sys{EntRec}), \textit{consistency score} (\sys{CnstScore}), and \textit{accuracy} between the \texttt{AWQ}, \texttt{GPTQ}, \texttt{bib8}, \texttt{bib4} quantized and full-precision models of \lm{Llama3-8B}, evaluated across all layers.}
  \label{fig:diff_llama}
\end{figure*}

\input{table/llama_lmhr}
Table~\ref{tab:lmhr_llama} shows the latent multi-hop reasoning accuracy comparison between full-precision models and quantized models. Additionally, Figure~\ref{fig:diff_7b}, Figure~\ref{fig:diff_14b}, and Figure~\ref{fig:diff_llama} display the differences in the \textit{entity recall score}, \textit{consistency score}, and \textit{accuracy} between the \texttt{AWQ}, \texttt{GPTQ8}, \texttt{GPTQ4}, \texttt{bib8}, \texttt{bib4} quantized and full-precision models of \lm{Qwen2.5-7B}, \lm{Qwen2.5-14B}, and \lm{Llama3-8B}, evaluated across all layers.

% --------------------------------------------------

% --------------------------------------------------

% --------------------------------------------------

% --------------------------------------------------

% --------------------------------------------------

% --------------------------------------------------

\input{table/factual_recall_details}

\end{document}

%% file: table/model.tex
\begin{table*}[t!]
    \centering
    \resizebox{\textwidth}{!}{%
        \begin{tabular}{rcccc}

        \toprule
        \textbf{Name}& \textbf{Citation} & \textbf{Size} & \textbf{Precision} & \textbf{Link}\\

        % \midrule

        % \lm{Vicuna} & \cite{chiang-lee-2023-large} & 13B & full & \url{https://huggingface.co/lmsys/vicuna-13b-v1.5}\\

        % \lm{Vicuna} & \cite{chiang-lee-2023-large} & 13B & GPTQ4 & \url{https://huggingface.co/TheBloke/vicuna-13B-v1.5-GPTQ}\\

        % \lm{Vicuna} & \cite{chiang-lee-2023-large} & 13B & GPTQ8 & \scriptsize{\url{https://huggingface.co/TheBloke/vicuna-13B-v1.5-GPTQ/tree/gptq-8bit--1g-actorder_True}}\\

        % \lm{Vicuna} & \cite{chiang-lee-2023-large} & 13B & AWQ & \url{https://huggingface.co/TheBloke/vicuna-13B-v1.5-AWQ}\\

        \midrule
        \lm{Llama3} & \citet{llama3modelcard} & 8B & full & \url{https://huggingface.co/meta-llama/Meta-Llama-3-8B-Instruct}\\

        \lm{Llama3} & \citet{llama3modelcard} & 8B & GPTQ4 & \url{https://huggingface.co/TechxGenus/Meta-Llama-3-8B-Instruct-GPTQ}\\

        \lm{Llama3} & \citet{llama3modelcard} & 8B & AWQ & \url{https://huggingface.co/TechxGenus/Meta-Llama-3-8B-Instruct-AWQ}\\

        % \midrule

        % \lm{Llama3} & \citet{llama3modelcard} & 70B & full & \url{https://huggingface.co/meta-llama/Meta-Llama-3-70B-Instruct}\\

        % \lm{Llama3} & \citet{llama3modelcard} & 70B & GPTQ4 & \url{https://huggingface.co/TechxGenus/Meta-Llama-3-70B-Instruct-GPTQ}\\

        % \lm{Llama3} & \citet{llama3modelcard} & 70B & AWQ & \url{https://huggingface.co/TechxGenus/Meta-Llama-3-70B-Instruct-AWQ}\\
        
        \midrule

        \lm{Qwen2.5} & \citet{qwen2024qwen25technicalreport} & 7B & Full & \url{https://huggingface.co/Qwen/Qwen2.5-7B-Instruct}\\
        \lm{Qwen2.5} & \citet{qwen2024qwen25technicalreport} & 7B & AWQ & \url{https://huggingface.co/Qwen/Qwen2.5-7B-Instruct-AWQ}\\
        \lm{Qwen2.5} & \citet{qwen2024qwen25technicalreport} & 7B & GPTQ4 & \url{https://huggingface.co/Qwen/Qwen2.5-7B-Instruct-GPTQ-Int4}\\
        \lm{Qwen2.5} & \citet{qwen2024qwen25technicalreport} & 7B & GPTQ8 & \url{https://huggingface.co/Qwen/Qwen2.5-7B-Instruct-GPTQ-Int8}\\

        \midrule
        
        \lm{Qwen2.5} & \citet{qwen2024qwen25technicalreport} & 14B & Full & \url{https://huggingface.co/Qwen/Qwen2.5-14B-Instruct}\\
        \lm{Qwen2.5} & \citet{qwen2024qwen25technicalreport} & 14B & AWQ & \url{https://huggingface.co/Qwen/Qwen2.5-14B-Instruct-AWQ}\\
        \lm{Qwen2.5} & \citet{qwen2024qwen25technicalreport} & 14B & GPTQ4 & \url{https://huggingface.co/Qwen/Qwen2.5-14B-Instruct-GPTQ-Int4}\\
        \lm{Qwen2.5} & \citet{qwen2024qwen25technicalreport} & 14B & GPTQ8 & \url{https://huggingface.co/Qwen/Qwen2.5-14B-Instruct-GPTQ-Int8}\\

        % \midrule
        
        % \lm{Qwen2.5} & \citet{qwen2024qwen25technicalreport} & 32B & Full & \url{https://huggingface.co/Qwen/Qwen2.5-32B-Instruct}\\
        % \lm{Qwen2.5} & \citet{qwen2024qwen25technicalreport} & 32B & AWQ & \url{https://huggingface.co/Qwen/Qwen2.5-32B-Instruct-AWQ}\\
        % \lm{Qwen2.5} & \citet{qwen2024qwen25technicalreport} & 32B & GPTQ4 & \url{https://huggingface.co/Qwen/Qwen2.5-32B-Instruct-GPTQ-Int4}\\
        % \lm{Qwen2.5} & \citet{qwen2024qwen25technicalreport} & 32B & GPTQ8 & \url{https://huggingface.co/Qwen/Qwen2.5-32B-Instruct-GPTQ-Int8}\\
        
        % \midrule
        % \lm{Qwen2.5} & \citet{qwen2024qwen25technicalreport} & 72B & Full & \url{https://huggingface.co/Qwen/Qwen2.5-72B-Instruct}\\
        % \lm{Qwen2.5} & \citet{qwen2024qwen25technicalreport} & 72B & AWQ & \url{https://huggingface.co/Qwen/Qwen2.5-72B-Instruct-AWQ}\\
        % \lm{Qwen2.5} & \citet{qwen2024qwen25technicalreport} & 72B & GPTQ4 & \url{https://huggingface.co/Qwen/Qwen2.5-72B-Instruct-GPTQ-Int4}\\
        % \lm{Qwen2.5} & \citet{qwen2024qwen25technicalreport} & 72B & GPTQ8 & \url{https://huggingface.co/Qwen/Qwen2.5-72B-Instruct-GPTQ-Int8}\\
        
        \bottomrule
        \end{tabular}
        }
    \caption{
    Detailed information about used LLMs in our experiments. 
    }
    \label{tab:used_model}
\end{table*}

%% file: table/factual_recall.tex
\begin{table}[t!]
    \centering
    \setlength{\extrarowheight}{3pt}
    \renewcommand*{\arraystretch}{0.5}
    
    \footnotesize
    \resizebox{\columnwidth}{!}{%
        \begin{tabular}{c|c|cc}
        \toprule[1.5pt]
        {\centering \scriptsize{\textbf{Model}}} 
        & Method & Accuracy \% $\uparrow$ & \# Correct\\
        \midrule
        \multirow{6}{*}{\vspace*{8pt}\centering \rotatebox[origin=r]{90}{\scriptsize{\lm{Qwen2.5-7B}}}} 
        & \cellcolor[HTML]{F2CFC2}full  & \cellcolor[HTML]{F2CFC2}\textbf{63.25} & \cellcolor[HTML]{F2CFC2}\textbf{6133} \\
        & bib4  & 60.72 & 5887  \\
        & bib8  & 63.01 & 6109  \\
        & gptq4 & 60.10 & 5827\\
        & gptq8 & \underline{63.22} & \underline{6130} \\
        & awq   & 60.60 & 5876 \\
        
        \midrule
        
        \multirow{6}{*}{\vspace*{8pt}\centering \rotatebox[origin=r]{90}{\scriptsize{\lm{Qwen2.5-14B}}}} 
        & \cellcolor[HTML]{F2CFC2}full  & \cellcolor[HTML]{F2CFC2}\textbf{73.08} & \cellcolor[HTML]{F2CFC2}\textbf{7086}\\
        & bib4  & 70.33 & 6819 \\
        & bib8  & \underline{73.06} & \underline{7084} \\
        & gptq4 & 25.20 & 2443 \\
        & gptq8 & 73.03 & 7081\\
        & awq   & 70.61 & 6846\\

        \midrule
        \multirow{5}{*}{\vspace*{8pt}\centering \rotatebox[origin=r]{90}{\scriptsize{\lm{Llama3-8B}}}} 
        & \cellcolor[HTML]{F2CFC2}full  & \cellcolor[HTML]{F2CFC2}\textbf{77.62} & \cellcolor[HTML]{F2CFC2}\textbf{7526}\\
        & bib4  & 72.19 & 7000\\
        & bib8  & \underline{76.95} & \underline{7461}\\
        & gptq8 & 71.39 & 6922\\
        & awq   & 71.83 & 6965\\
        
        \toprule[1.5pt]
        \end{tabular}
    }
    \caption{Knowledge recall accuracy results (\%) and number of correct predictions (out of 9696 queries) on the \data{LRE} dataset for \texttt{Qwen2.5-\{7B,14B\}} and \texttt{Llama3-8B} models across different quantization methods.}
    \label{tab:lre_acc}
\end{table}

%% file: table/llama_lmhr.tex
\begin{table}[t!]
    \centering
    \renewcommand*{\arraystretch}{0.5}
    
    \footnotesize
    \resizebox{\columnwidth}{!}{%
        \begin{tabular}{c|c|cc|c}
        \toprule[1.5pt]
        {\centering \scriptsize{\textbf{Model}}} 
        & Method & $r_1(e_1)$ $\uparrow$ & $r_2(e_2)$ $\uparrow$ & $r_2(r_1(e_1))$ $\uparrow$ \\
        \midrule

         \multirow{6}{*}{\centering \rotatebox[origin=c]{90}{\scriptsize{\lm{Qwen2.5-7B}}}} 
        & \cellcolor[HTML]{F2CFC2}{full}  & \cellcolor[HTML]{F2CFC2}\textbf{25.03} & \cellcolor[HTML]{F2CFC2}39.07 & \cellcolor[HTML]{F2CFC2}\textbf{20.61} \\
        % \cdashline{2-5}
        & bib4  & 25.01 & 39.02 & 20.61  \\
        & bib8  & \underline{25.01} & 39.02 & \underline{20.61}\\
        & gptq4 & 22.29 & 37.44 & 19.59  \\
        & gptq8 & 24.76 & \textbf{\underline{39.15}} & 20.55  \\
        & awq   & 22.07 & 38.89 & 18.05   \\

        \midrule

        \multirow{6}{*}{\centering \rotatebox[origin=c]{90}{\scriptsize{\lm{Qwen2.5-14B}}}} 
        & \cellcolor[HTML]{F2CFC2}full  & \cellcolor[HTML]{F2CFC2}\textbf{35.23} & \cellcolor[HTML]{F2CFC2}40.45 & \cellcolor[HTML]{F2CFC2}24.72  \\
        % \cdashline{2-5}
        & bib4  & 35.16 & 40.56 & 24.76  \\
        & bib8  & \underline{35.16} & 40.56 & \textbf{\underline{24.76}}  \\
        & gptq4 & 32.22 & \textbf{\underline{45.19}} & 23.29   \\
        & gptq8 & 35.16 & 39.55 & 24.61   \\
        & awq   & 24.69 & 35.61 & 21.80   \\

        \midrule
        
        \multirow{5}{*}{\centering \rotatebox[origin=c]{90}{\scriptsize{\lm{Llama3-8B}}}} 
        & \cellcolor[HTML]{F2CFC2}{full}  & \cellcolor[HTML]{F2CFC2}7.79 & \cellcolor[HTML]{F2CFC2}21.39 & \cellcolor[HTML]{F2CFC2}4.45 \\
        % \cdashline{2-5}
        & bib4  & 4.94 & 27.44 & 3.14 \\
        & bib8  & \underline{5.79} & \underline{18.27} & \underline{2.90}\\
        % & gptq4 & 22.29 & 37.44 & 19.59  \\
        & gptq8 & \textbf{23.62} & \textbf{40.73} & \textbf{20.94}  \\
        & awq   & 22.35 & 37.79 & 19.56   \\

        % \midrule

        % \multirow{6}{*}{\centering \rotatebox[origin=c]{90}{\scriptsize{\lm{Llama3-70B}}}} 
        % & \cellcolor[HTML]{F2CFC2}{full} & \cellcolor[HTML]{F2CFC2} & \cellcolor[HTML]{F2CFC2} & \cellcolor[HTML]{F2CFC2} \\
        %  & bib4  & 25.01 & 39.02 & 20.61  \\
        % & bib8  & \underline{25.01} & 39.02 & \underline{20.61}\\
        % & gptq4 & 22.29 & 37.44 & 19.59  \\
        % & gptq8 & 24.76 & \textbf{\underline{39.15}} & 20.55  \\
        % & awq   & 22.07 & 38.89 & 18.05   \\

        \toprule[1.5pt]
        \end{tabular}
    }
    \caption{Accuracy (in \%) of different models in connecting and traversing implicit knowledge to successfully answer latent multi-hop queries on \data{TwoHop-Fact}.}
    \label{tab:lmhr_llama}
\end{table}

%% file: table/factual_recall_details.tex
\begin{table*}[h]
  \centering
  % \footnotesize
    \begin{tabular}{l|cccccc}
    \toprule
    \lm{Qwen2.5-7B} & full  & bib4  & bib8  & gptq4 & gptq8 & awq \\
    
    \midrule
    city in country              & 93.00 & 85.00 & 93.00 & 93.00 & 93.00 & 93.00 \\
    company CEO                  & 49.00 & 39.00 & 50.00 & 45.00 & 50.00 & 46.00 \\
    company hq                   & 59.00 & 54.00 & 59.00 & 56.00 & 58.00 & 59.00 \\
    country capital city         & 100.00 & 100.00 & 100.00 & 100.00 & 100.00 & 100.00 \\
    country currency             & 93.00 & 90.00 & 93.00 & 87.00 & 93.00 & 97.00 \\
    country language             & 92.00 & 88.00 & 92.00 & 88.00 & 92.00 & 92.00 \\
    country largest city         & 100.00 & 100.00 & 100.00 & 100.00 & 100.00 & 100.00 \\
    food from country            & 83.00 & 83.00 & 83.00 & 80.00 & 83.00 & 80.00 \\
    landmark in country          & 81.00 & 80.00 & 81.00 & 79.00 & 81.00 & 80.00 \\
    landmark on continent & 87.00 & 84.00 & 85.00 & 83.00 & 87.00 & 85.00 \\
    person father       & 45.00 & 39.00 & 45.00 & 40.00 & 45.00 & 41.00 \\
    person lead singer of band   & 100.00 & 100.00 & 100.00 & 100.00 & 100.00 & 100.00 \\
    person mother                & 39.00 & 32.00 & 39.00 & 34.00 & 38.00 & 33.00 \\
    person native language       & 90.00 & 91.00 & 90.00 & 88.00 & 90.00 & 86.00 \\
    person occupation            & 33.00 & 34.00 & 33.00 & 31.00 & 33.00 & 30.00 \\
    person plays instrument      & 44.00 & 46.00 & 43.00 & 42.00 & 44.00 & 43.00 \\
    person sport position & 66.00 & 63.00 & 66.00 & 59.00 & 66.00 & 62.00 \\
    person university            & 47.00 & 46.00 & 46.00 & 48.00 & 46.00 & 47.00 \\
    plays pro sport              & 79.00 & 77.00 & 81.00 & 75.00 & 79.00 & 76.00 \\
    pokemon evolution            & 100.00 & 89.00 & 100.00 & 91.00 & 100.00 & 95.00 \\
    president birth year         & 0.00 & 0.00 & 0.00 & 0.00 & 0.00 & 0.00 \\
    president election year      & 0.00 & 0.00 & 0.00 & 0.00 & 0.00 & 0.00 \\
    product by company           & 80.00 & 77.00 & 79.00 & 77.00 & 80.00 & 79.00 \\
    star constellation name      & 85.00 & 85.00 & 85.00 & 85.00 & 85.00 & 84.00 \\
    superhero archnemesis        & 34.00 & 33.00 & 33.00 & 33.00 & 34.00 & 31.00 \\
    superhero person             & 49.00 & 50.00 & 50.00 & 49.00 & 50.00 & 48.00 \\
    AVG                          & 63.25 & 60.72 & 63.01 & 60.10 & 63.22 & 60.00 \\
    \bottomrule
    \end{tabular}%
  \caption{Per-relation knowledge recall accuracy results (\%) on the \data{LRE} dataset for \lm{Qwen2.5-7B} across different quantization methods.}
  \label{tab:per-relation-factual-recall-acc-7B}%
\end{table*}%

\begin{table*}[h]
  \centering
  % \footnotesize
    \begin{tabular}{l|cccccc}
    \toprule
    \lm{Qwen2.5-14B} & full  & bib4  & bib8  & gptq4 & gptq8 & awq \\
    \midrule
    city in country              & 96.00 & 96.00 & 96.00 & 81.00 & 96.00 & 96.00 \\
    company CEO                  & 58.00 & 54.00 & 59.00 & 2.00  & 58.00 & 57.00 \\
    company hq                   & 68.00 & 65.00 & 66.00 & 10.00 & 69.00 & 65.00 \\
    country capital city         & 100.00 & 100.00 & 100.00 & 79.00 & 100.00 & 100.00 \\
    country currency             & 90.00 & 83.00 & 87.00 & 77.00 & 87.00 & 90.00 \\
    country language             & 96.00 & 96.00 & 96.00 & 92.00 & 96.00 & 96.00 \\
    country largest city         & 96.00 & 100.00 & 96.00 & 83.00 & 96.00 & 96.00 \\
    food from country            & 80.00 & 77.00 & 80.00 & 50.00 & 80.00 & 80.00 \\
    landmark in country          & 87.00 & 86.00 & 87.00 & 31.00 & 87.00 & 87.00 \\
    landmark on continent        & 85.00 & 88.00 & 84.00 & 48.00 & 85.00 & 84.00 \\
    person father                & 63.00 & 56.00 & 62.00 & 7.00  & 63.00 & 57.00 \\
    person lead singer of band   & 100.00 & 100.00 & 100.00 & 29.00 & 100.00 & 100.00 \\
    person mother                & 59.00 & 49.00 & 59.00 & 1.00  & 59.00 & 53.00 \\
    person native language       & 94.00 & 94.00 & 94.00 & 39.00 & 93.00 & 93.00 \\
    person occupation            & 47.00 & 45.00 & 49.00 & 12.00 & 49.00 & 45.00 \\
    person plays instrument      & 67.00 & 61.00 & 68.00 & 52.00 & 66.00 & 63.00 \\
    person sport position        & 76.00 & 74.00 & 76.00 & 5.00  & 74.00 & 71.00 \\
    person university            & 48.00 & 45.00 & 52.00 & 43.00 & 47.00 & 46.00 \\
    plays pro sport              & 83.00 & 83.00 & 81.00 & 48.00 & 82.00 & 85.00 \\
    pokemon evolution            & 100.00 & 100.00 & 100.00 & 30.00 & 100.00 & 100.00 \\
    president birth year         & 0.00  & 0.00  & 0.00  & 0.00  & 0.00  & 0.00 \\
    president election year      & 0.00  & 0.00  & 0.00  & 0.00  & 0.00  & 0.00 \\
    product by company           & 85.00 & 82.00 & 85.00 & 38.00 & 85.00 & 84.00 \\
    star constellation name      & 87.00 & 87.00 & 87.00 & 75.00 & 87.00 & 87.00 \\
    superhero archnemesis        & 48.00 & 43.00 & 46.00 & 4.00  & 48.00 & 44.00 \\
    superhero person             & 76.00 & 73.00 & 75.00 & 5.00  & 77.00 & 72.00 \\
    AVG                          & 73.08 & 70.33 & 73.06 & 25.20 & 73.03 & 70.61 \\
    \bottomrule
    \end{tabular}%
  \caption{Per-relation knowledge recall accuracy results (\%) on the \data{LRE} dataset for \lm{Qwen2.5-14B} across different quantization methods.}
  \label{tab:per-relation-factual-recall-acc-14B}%
\end{table*}%

%% file: custom.bib
@inproceedings{choe-etal-2025-autoregressive,
    title = "Do All Autoregressive Transformers Remember Facts the Same Way? A Cross-Architecture Analysis of Recall Mechanisms",
    author = "Choe, Minyeong  and
      Cho, Haehyun  and
      Seo, Changho  and
      Kim, Hyunil",
    editor = "Christodoulopoulos, Christos  and
      Chakraborty, Tanmoy  and
      Rose, Carolyn  and
      Peng, Violet",
    booktitle = "Proceedings of the 2025 Conference on Empirical Methods in Natural Language Processing",
    month = nov,
    year = "2025",
    address = "Suzhou, China",
    publisher = "Association for Computational Linguistics",
    url = "https://aclanthology.org/2025.emnlp-main.1448/",
    doi = "10.18653/v1/2025.emnlp-main.1448",
    pages = "28494--28513",
    ISBN = "979-8-89176-332-6",
}

@inproceedings{dettmers-etal-2022-8bits,
    author = {Dettmers, Tim and Lewis, Mike and Belkada, Younes and Zettlemoyer, Luke},
    title = {LLM.int8(): 8-bit matrix multiplication for transformers at scale},
    year = {2022},
    isbn = {9781713871088},
    publisher = {Curran Associates Inc.},
    address = {Red Hook, NY, USA},
    booktitle = {Proceedings of the 36th International Conference on Neural Information Processing Systems},
    articleno = {2198},
    numpages = {15},
    location = {New Orleans, LA, USA},
    series = {NIPS '22}
}

@inproceedings{
    frantar-etal-2023-optq,
    title={{OPTQ}: Accurate Quantization for Generative Pre-trained Transformers},
    author={Elias Frantar and Saleh Ashkboos and Torsten Hoefler and Dan Alistarh},
    booktitle={The Eleventh International Conference on Learning Representations },
    year={2023},
    url={https://openreview.net/forum?id=tcbBPnfwxS}
}

@inproceedings{goncalves-strubell-2023-understanding,
    title = "Understanding the Effect of Model Compression on Social Bias in Large Language Models",
    author = "Gon{\c{c}}alves, Gustavo  and
      Strubell, Emma",
    editor = "Bouamor, Houda  and
      Pino, Juan  and
      Bali, Kalika",
    booktitle = "Proceedings of the 2023 Conference on Empirical Methods in Natural Language Processing",
    month = dec,
    year = "2023",
    address = "Singapore",
    publisher = "Association for Computational Linguistics",
    url = "https://aclanthology.org/2023.emnlp-main.161",
    doi = "10.18653/v1/2023.emnlp-main.161",
    pages = "2663--2675"
}

@inproceedings{hernandez-2024-linearity,
    title={Linearity of Relation Decoding in Transformer Language Models},
    author={Evan Hernandez and Arnab Sen Sharma and Tal Haklay and Kevin Meng and Martin Wattenberg and Jacob Andreas and Yonatan Belinkov and David Bau},
    booktitle={The Twelfth International Conference on Learning Representations},
    year={2024},
    url={https://openreview.net/forum?id=w7LU2s14kE}
}

@InProceedings{hong-etal-2024-trust,
  title = 	 {Decoding Compressed Trust: Scrutinizing the Trustworthiness of Efficient {LLM}s Under Compression},
  author =       {Hong, Junyuan and Duan, Jinhao and Zhang, Chenhui and Li, Zhangheng and Xie, Chulin and Lieberman, Kelsey and Diffenderfer, James and Bartoldson, Brian R. and Jaiswal, Ajay Kumar and Xu, Kaidi and Kailkhura, Bhavya and Hendrycks, Dan and Song, Dawn and Wang, Zhangyang and Li, Bo},
  booktitle = 	 {Proceedings of the 41st International Conference on Machine Learning},
  pages = 	 {18611--18633},
  year = 	 {2024},
  editor = 	 {Salakhutdinov, Ruslan and Kolter, Zico and Heller, Katherine and Weller, Adrian and Oliver, Nuria and Scarlett, Jonathan and Berkenkamp, Felix},
  volume = 	 {235},
  series = 	 {Proceedings of Machine Learning Research},
  month = 	 {21--27 Jul},
  publisher =    {PMLR},
  url = 	 {https://proceedings.mlr.press/v235/hong24a.html},
}

@inproceedings{jin-etal-2024-comprehensive,
    title = "A Comprehensive Evaluation of Quantization Strategies for Large Language Models",
    author = "Jin, Renren  and
      Du, Jiangcun  and
      Huang, Wuwei  and
      Liu, Wei  and
      Luan, Jian  and
      Wang, Bin  and
      Xiong, Deyi",
    editor = "Ku, Lun-Wei  and
      Martins, Andre  and
      Srikumar, Vivek",
    booktitle = "Findings of the Association for Computational Linguistics: ACL 2024",
    month = aug,
    year = "2024",
    address = "Bangkok, Thailand",
    publisher = "Association for Computational Linguistics",
    url = "https://aclanthology.org/2024.findings-acl.726",
    doi = "10.18653/v1/2024.findings-acl.726",
    pages = "12186--12215"
}

@ARTICLE{gray-etal-1998-quantization,
  author={Gray, R.M. and Neuhoff, D.L.},
  journal={IEEE Transactions on Information Theory}, 
  title={Quantization}, 
  year={1998},
  volume={44},
  number={6},
  pages={2325-2383},
  doi={10.1109/18.720541}
}

@inproceedings{
    kirsten-etal-2024-bias,
    title = "The Impact of Inference Acceleration on Bias of {LLM}s",
    author = "Kirsten, Elisabeth  and
      Habernal, Ivan  and
      Nanda, Vedant  and
      Zafar, Muhammad Bilal",
    editor = "Chiruzzo, Luis  and
      Ritter, Alan  and
      Wang, Lu",
    booktitle = "Proceedings of the 2025 Conference of the Nations of the Americas Chapter of the Association for Computational Linguistics: Human Language Technologies (Volume 1: Long Papers)",
    month = apr,
    year = "2025",
    address = "Albuquerque, New Mexico",
    publisher = "Association for Computational Linguistics",
    url = "https://aclanthology.org/2025.naacl-long.91/",
    doi = "10.18653/v1/2025.naacl-long.91",
    pages = "1834--1853",
    ISBN = "979-8-89176-189-6",
}

@inproceedings{
    li2024investigating,
    title={Investigating the Impact of Quantization on Adversarial Robustness},
    author={Qun Li and Yuan Meng and Chen Tang and Jiacheng Jiang and Zhi Wang},
    booktitle={5th Workshop on practical ML for limited/low resource settings},
    year={2024},
    url={https://openreview.net/forum?id=TQnw5RIeK2}
}

@inproceedings{lin-etal-2024-awq,
     author = {Lin, Ji and Tang, Jiaming and Tang, Haotian and Yang, Shang and Chen, Wei-Ming and Wang, Wei-Chen and Xiao, Guangxuan and Dang, Xingyu and Gan, Chuang and Han, Song},
     booktitle = {Proceedings of Machine Learning and Systems},
     editor = {P. Gibbons and G. Pekhimenko and C. De Sa},
     pages = {87--100},
     title = {AWQ: Activation-aware Weight Quantization for On-Device LLM Compression and Acceleration},
     url = {https://proceedings.mlsys.org/paper_files/paper/2024/file/42a452cbafa9dd64e9ba4aa95cc1ef21-Paper-Conference.pdf},
     volume = {6},
     year = {2024}
}

@inproceedings{liu-etal-2024-emergent,
    title = "Do Emergent Abilities Exist in Quantized Large Language Models: An Empirical Study",
    author = "Liu, Peiyu  and
      Liu, Zikang  and
      Gao, Ze-Feng  and
      Gao, Dawei  and
      Zhao, Wayne Xin  and
      Li, Yaliang  and
      Ding, Bolin  and
      Wen, Ji-Rong",
    editor = "Calzolari, Nicoletta  and
      Kan, Min-Yen  and
      Hoste, Veronique  and
      Lenci, Alessandro  and
      Sakti, Sakriani  and
      Xue, Nianwen",
    booktitle = "Proceedings of the 2024 Joint International Conference on Computational Linguistics, Language Resources and Evaluation (LREC-COLING 2024)",
    month = may,
    year = "2024",
    address = "Torino, Italia",
    publisher = "ELRA and ICCL",
    url = "https://aclanthology.org/2024.lrec-main.461",
    pages = "5174--5190"
}

@article{llama3modelcard,
    title={Llama 3 Model Card},
    author={AI@Meta},
    year={2024},
    url = {https://github.com/meta-llama/llama3/blob/main/MODEL_CARD.md}
}

@inproceedings{marchisio-etal-2024-quantization,
    title = "How Does Quantization Affect Multilingual {LLM}s?",
    author = {Marchisio, Kelly  and
      Dash, Saurabh  and
      Chen, Hongyu  and
      Aumiller, Dennis  and
      {\"U}st{\"u}n, Ahmet  and
      Hooker, Sara  and
      Ruder, Sebastian},
    editor = "Al-Onaizan, Yaser  and
      Bansal, Mohit  and
      Chen, Yun-Nung",
    booktitle = "Findings of the Association for Computational Linguistics: EMNLP 2024",
    month = nov,
    year = "2024",
    address = "Miami, Florida, USA",
    publisher = "Association for Computational Linguistics",
    url = "https://aclanthology.org/2024.findings-emnlp.935",
    pages = "15928--15947"
}

@inproceedings{namburi-etal-2023-cost,
    title = "The Cost of Compression: Investigating the Impact of Compression on Parametric Knowledge in Language Models",
    author = "Namburi, Satya Sai Srinath  and
      Sreedhar, Makesh  and
      Srinivasan, Srinath  and
      Sala, Frederic",
    editor = "Bouamor, Houda  and
      Pino, Juan  and
      Bali, Kalika",
    booktitle = "Findings of the Association for Computational Linguistics: EMNLP 2023",
    month = dec,
    year = "2023",
    address = "Singapore",
    publisher = "Association for Computational Linguistics",
    url = "https://aclanthology.org/2023.findings-emnlp.349/",
    doi = "10.18653/v1/2023.findings-emnlp.349",
    pages = "5255--5273"
}

@InProceedings{park-etal-2022-robust,
    author="Park, Sein
    and Jang, Yeongsang
    and Park, Eunhyeok",
    editor="Avidan, Shai
    and Brostow, Gabriel
    and Ciss{\'e}, Moustapha
    and Farinella, Giovanni Maria
    and Hassner, Tal",
    title="Symmetry Regularization and Saturating Nonlinearity for Robust Quantization",
    booktitle="Computer Vision -- ECCV 2022",
    year="2022",
    publisher="Springer Nature Switzerland",
    address="Cham",
    pages="206--222",
    isbn="978-3-031-20083-0"
}

@misc{qwen2024qwen25technicalreport,
      title={Qwen2.5 Technical Report}, 
      author={Qwen and An Yang and Baosong Yang and Beichen Zhang and Binyuan Hui and Bo Zheng and Bowen Yu and Chengyuan Li and Dayiheng Liu and Fei Huang and Haoran Wei and Huan Lin and Jian Yang and Jianhong Tu and Jianwei Zhang and Jianxin Yang and Jiaxi Yang and Jingren Zhou and Junyang Lin and Kai Dang and Keming Lu and Keqin Bao and Kexin Yang and Le Yu and Mei Li and Mingfeng Xue and Pei Zhang and Qin Zhu and Rui Men and Runji Lin and Tianhao Li and Tingyu Xia and Xingzhang Ren and Xuancheng Ren and Yang Fan and Yang Su and Yichang Zhang and Yu Wan and Yuqiong Liu and Zeyu Cui and Zhenru Zhang and Zihan Qiu},
      year={2024},
      eprint={2412.15115},
      archivePrefix={arXiv},
      primaryClass={cs.CL},
      url={https://arxiv.org/abs/2412.15115}, 
}

@inproceedings{ramesh-etal-2023-comparative,
    title = "A Comparative Study on the Impact of Model Compression Techniques on Fairness in Language Models",
    author = "Ramesh, Krithika  and
      Chavan, Arnav  and
      Pandit, Shrey  and
      Sitaram, Sunayana",
    editor = "Rogers, Anna  and
      Boyd-Graber, Jordan  and
      Okazaki, Naoaki",
    booktitle = "Proceedings of the 61st Annual Meeting of the Association for Computational Linguistics (Volume 1: Long Papers)",
    month = jul,
    year = "2023",
    address = "Toronto, Canada",
    publisher = "Association for Computational Linguistics",
    url = "https://aclanthology.org/2023.acl-long.878",
    doi = "10.18653/v1/2023.acl-long.878",
    pages = "15762--15782"
}

@inproceedings{singh-sajjad-2025-interpreting,
    title = "Interpreting the Effects of Quantization on {LLM}s",
    author = "Singh, Manpreet  and
      Sajjad, Hassan",
    editor = "Inui, Kentaro  and
      Sakti, Sakriani  and
      Wang, Haofen  and
      Wong, Derek F.  and
      Bhattacharyya, Pushpak  and
      Banerjee, Biplab  and
      Ekbal, Asif  and
      Chakraborty, Tanmoy  and
      Singh, Dhirendra Pratap",
    booktitle = "Proceedings of the 14th International Joint Conference on Natural Language Processing and the 4th Conference of the Asia-Pacific Chapter of the Association for Computational Linguistics",
    month = dec,
    year = "2025",
    address = "Mumbai, India",
    publisher = "The Asian Federation of Natural Language Processing and The Association for Computational Linguistics",
    url = "https://aclanthology.org/2025.ijcnlp-long.123/",
    pages = "2267--2281",
    ISBN = "979-8-89176-298-5",
}

@misc{wang2026largelanguagemodelsexplain,
      title={Can Large Language Models Still Explain Themselves? Investigating the Impact of Quantization on Self-Explanations}, 
      author={Qianli Wang and Nils Feldhus and Pepa Atanasova and Fedor Splitt and Simon Ostermann and Sebastian Möller and Vera Schmitt},
      year={2026},
      eprint={2601.00282},
      archivePrefix={arXiv},
      primaryClass={cs.CL},
      url={https://arxiv.org/abs/2601.00282}, 
}

@inproceedings{xiao-etal-2023-smooth,
    author = {Xiao, Guangxuan and Lin, Ji and Seznec, Mickael and Wu, Hao and Demouth, Julien and Han, Song},
    title = {SmoothQuant: accurate and efficient post-training quantization for large language models},
    year = {2023},
    publisher = {JMLR.org},
    booktitle = {Proceedings of the 40th International Conference on Machine Learning},
    articleno = {1585},
    numpages = {13},
    location = {Honolulu, Hawaii, USA},
    series = {ICML'23}
}

@inproceedings{yang-2024-latent-multi-hop-reasoning,
    title = "Do Large Language Models Latently Perform Multi-Hop Reasoning?",
    author = "Yang, Sohee  and
      Gribovskaya, Elena  and
      Kassner, Nora  and
      Geva, Mor  and
      Riedel, Sebastian",
    editor = "Ku, Lun-Wei  and
      Martins, Andre  and
      Srikumar, Vivek",
    booktitle = "Proceedings of the 62nd Annual Meeting of the Association for Computational Linguistics (Volume 1: Long Papers)",
    month = aug,
    year = "2024",
    address = "Bangkok, Thailand",
    publisher = "Association for Computational Linguistics",
    url = "https://aclanthology.org/2024.acl-long.550/",
    doi = "10.18653/v1/2024.acl-long.550",
    pages = "10210--10229"
}

@inproceedings{yu-ananiadou-2024-neuron,
    title = "Neuron-Level Knowledge Attribution in Large Language Models",
    author = "Yu, Zeping  and
      Ananiadou, Sophia",
    editor = "Al-Onaizan, Yaser  and
      Bansal, Mohit  and
      Chen, Yun-Nung",
    booktitle = "Proceedings of the 2024 Conference on Empirical Methods in Natural Language Processing",
    month = nov,
    year = "2024",
    address = "Miami, Florida, USA",
    publisher = "Association for Computational Linguistics",
    url = "https://aclanthology.org/2024.emnlp-main.191/",
    doi = "10.18653/v1/2024.emnlp-main.191",
    pages = "3267--3280"
}

@article{zhu-2024-surveymodelcompressionlarge,
    author = {Zhu, Xunyu and Li, Jian and Liu, Yong and Ma, Can and Wang, Weiping},
    title = {A Survey on Model Compression for Large Language Models},
    journal = {Transactions of the Association for Computational Linguistics},
    volume = {12},
    pages = {1556-1577},
    year = {2024},
    month = {11},
    issn = {2307-387X},
    doi = {10.1162/tacl_a_00704},
    url = {https://doi.org/10.1162/tacl\_a\_00704},
    eprint = {https://direct.mit.edu/tacl/article-pdf/doi/10.1162/tacl\_a\_00704/2482209/tacl\_a\_00704.pdf},
}
